\documentclass[10pt,twocolumn,letterpaper]{article}

\usepackage[pagenumbers]{iccv} 

\usepackage{enumitem}
\usepackage{xspace}
\usepackage{booktabs}
\usepackage{colortbl}
\usepackage{pifont}
\usepackage{graphicx}
\usepackage{siunitx}
\usepackage{multirow}
\usepackage{amsmath}
\usepackage{amssymb}
\usepackage{bm}
\usepackage{soul}
\usepackage{comment}
\usepackage{epigraph}
\usepackage{algorithm}
\usepackage{algpseudocode}
\usepackage{tcolorbox}
\usepackage{sidecap}

\usepackage{caption}
\definecolor{Gray}{gray}{0.90}

\newcommand{\cmark}{\ding{51}}%
\newcommand{\xmark}{\ding{55}}%

\def\rvf{{\mathbf{f}}}

\def\rmF{{\mathbf{F}}}

\def\rmX{{\mathbf{X}}}
\def\rmY{{\mathbf{Y}}}

\DeclareMathAlphabet{\mathsfit}{\encodingdefault}{\sfdefault}{m}{sl}
\SetMathAlphabet{\mathsfit}{bold}{\encodingdefault}{\sfdefault}{bx}{n}

\def\gD{{\mathcal{D}}}
\def\gE{{\mathcal{E}}}

\def\gG{{\mathcal{G}}}

\def\gL{{\mathcal{L}}}

\def\gN{{\mathcal{N}}}

\def\sA{{\mathbb{A}}}

\def\sF{{\mathbb{F}}}

\def\sR{{\mathbb{R}}}

\def\sY{{\mathbb{Y}}}

\def\pr{\text{P}}

\def\modelname{Seg-TTO\xspace}

\newcommand{\inc}[1]{\ensuremath{_{\text{\textcolor{PineGreen}{(+#1)}}}}}
\newcommand{\dec}[1]{\ensuremath{_{\text{\textcolor{RedOrange}{(-#1)}}}}}

\definecolor{iccvblue}{rgb}{0.21,0.49,0.74}
\usepackage[pagebackref,breaklinks,colorlinks,allcolors=iccvblue]{hyperref}

\def\paperID{11152} 
\def\confName{ICCV}
\def\confYear{2025}

\title{Test-Time Optimization for Domain Adaptive Open Vocabulary Segmentation}

\author{
Ulindu De Silva\textsuperscript{1}\thanks{Equal Contribution} \qquad Didula Samaraweera\textsuperscript{1}\footnotemark[1] \qquad Sasini Wanigathunga\textsuperscript{1}\footnotemark[1] \qquad
Kavindu Kariyawasam\textsuperscript{1}\footnotemark[1] \\ 
Kanchana Ranasinghe\textsuperscript{2} \qquad Muzammal Naseer\textsuperscript{3} \qquad Ranga Rodrigo\textsuperscript{1} \\[1.5em]
\textsuperscript{1}University of Moratuwa \qquad \textsuperscript{2}Stony Brook University \qquad \textsuperscript{3}Khalifa University
\vspace{0.5em} \\ 
}

\begin{document}
\maketitle

\begin{abstract}
     We present \modelname, a novel framework for zero-shot, open-vocabulary semantic segmentation (OVSS), designed to excel in specialized domain tasks. While current open-vocabulary approaches show impressive performance on standard segmentation benchmarks under zero-shot settings, they fall short of supervised counterparts on highly domain-specific datasets. We focus on segmentation-specific test-time optimization to address this gap. 
    Segmentation requires an understanding of multiple concepts within a single image while retaining the locality and spatial structure of representations. We propose a novel self-supervised objective adhering to these requirements and use it to align the model parameters with input images at test time. In the textual modality, we learn multiple embeddings for each category to capture diverse concepts within an image, while in the visual modality, we calculate pixel-level losses followed by embedding aggregation operations specific to preserving spatial structure. 
    Our resulting framework termed \modelname is a plug-and-play module. We integrate \modelname with three state-of-the-art OVSS approaches and evaluate across 22 challenging OVSS tasks covering a range of specialized domains. Our \modelname demonstrates clear performance improvements (up to 27\% mIoU increase on some datasets) establishing new state-of-the-art. 
    Our code and models will be released publicly. 
\end{abstract}

\section{Introduction}

\label{sec:intro}

Open vocabulary semantic segmentation (OVSS) involves classifying each pixel of an image into an arbitrary number of categories given in the form of natural language. Recent works leverage contrastive vision-language models (VLMs) \cite{clip,align} to construct powerful OVSS models \cite{catseg,ovseg,xu2023side,clippy,clip-dinoiser,proxyclip} that can segment wide ranges of natural images under zero-shot settings. 
However, these models struggle in highly domain-specific tasks (e.g., medical, engineering, agriculture) performing subpar to their supervised counterparts \cite{MESS}. The nature of such tasks makes fully supervised approaches additionally expensive (e.g., only highly specialized individuals could annotate certain medical domain images). This underscores the importance of OVSS approaches that can accurately tackle these tasks in zero-shot settings.

\begin{figure}[t]
    \centering
    \includegraphics[width=\linewidth]{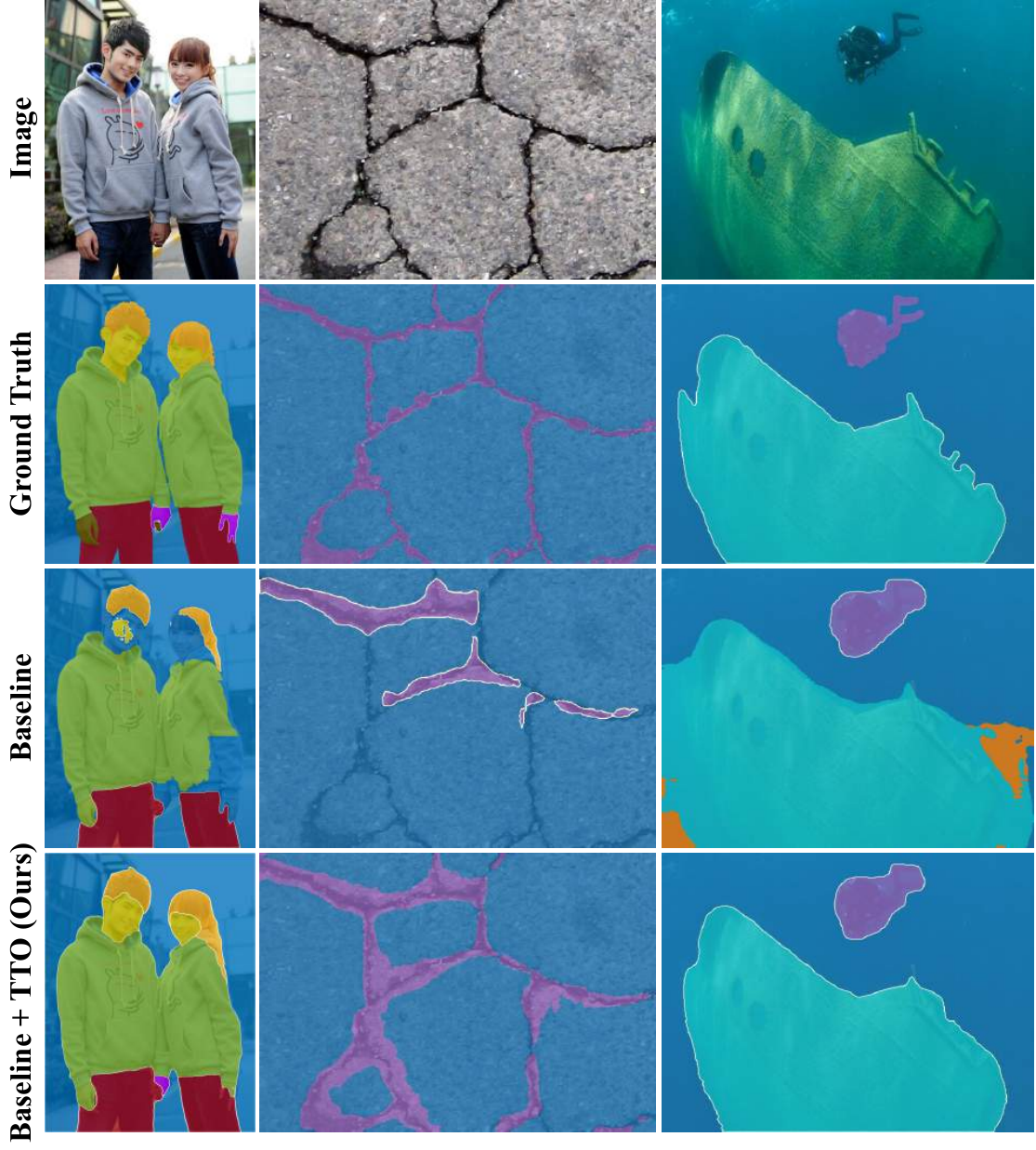}
    \caption{
    Our \modelname (row 4) improves state-of-the-art baseline CAT-Seg from \cite{catseg} (row 3) by segmenting missed regions as well as correcting incorrectly assigned labels. We attribute these improvements to the visual \& textual augmentations and the novel segmentation-specific test-time optimization used in our \modelname.
    }
    \label{fig:teaser}
\end{figure}

These tasks often involve drastic shifts across both visual and textual modalities such as images being captured from electromagnetic or multi-spectral sources, or category names being scientific or technical. We attribute the gap between zero-shot and supervised methods in these domains to such factors. Zero-shot approaches build off VLMs that may be unfamiliar to such out-of-domain concepts. In open-vocabulary classification, several recent works bridge this gap while retaining zero-shot ability through various test-time optimization strategies \cite{menon2022visual,ozturk2024intelligent,abdul2024align, MUST, llmmutate, tpt}. However, classification involves a single distinct category (or concept) per image that needs to be recognized. In contrast, segmentation can involve multiple categories per image, where each pixel must be classified into those distinct categories (\Cref{fig:teaser}), limiting the direct applicability of these ideas to OVSS tasks. In fact, test-time optimization for OVSS remains relatively unexplored. 

Motivated by these findings, we propose a test time optimization framework for OVSS. Segmentation tasks involving specialized domains (e.g., earth monitoring, medical sciences, or agriculture and biology) require an understanding of the novel categories in the language modality, with an emphasis on generating multi-category, pixel-level outputs. 
This requires visual features to preserve locality and spatial structure as illustrated in \Cref{fig:teaser}. Considering for example the left column in \Cref{fig:teaser}, the visual features of the blue category must avoid affecting the nearby surrounding features. Breaking the locality and structure could lead to incorrect predictions (e.g., row 3 in \Cref{fig:teaser}). 

Thus, while adopting pre-trained features for a given sample, we use specialized loss functions, learnable embeddings, and feature aggregation to preserve this spatial structure and separation of distinct concepts.
We propose a self-supervised objective to measure representation suitability for OVSS tasks. Our objective calculates cross-modal feature similarity and estimates suitability as a combination of feature entropy and pseudo-label-based cross-entropy measurements. We calculate pixel-level losses followed by locality-aware visual feature aggregation to retain spatial structure and per-category text embedding updates to better separate distinct concept features.

Revisiting the nature of specialized domain tasks, we note how pretrained features may be unfamiliar with certain concepts (e.g., ``mediastinum" in \Cref{fig:domain_analysis}).
Therein, we further augment text features with category descriptions describing distinct visual attributes. We use large language models (LLMs) known to contain extensive world knowledge \cite{Yu2023KoLACB} to generate these category attribute descriptions. 
At test time, we filter these attributes using similarity metrics in our model latent space conditioned on the test-time sample. This provides text representations that are distinct from other categories and relevant to the test-time sample.   

We then use these modified representations to generate segmentations for OVSS tasks entirely under zero-shot settings. 
We name our resulting framework as \modelname. 

\vspace{0.5em} \noindent
We summarize our key contributions as follows:
\begin{itemize}[leftmargin=3em,noitemsep,topsep=0.0ex,itemsep=-1.0ex,partopsep=0ex,parsep=1ex]
    \item First test-time optimization framework for OVSS operating zero-shot on specialized-domain tasks.   
    \item Novel prompt tuning strategy with losses suitable for dense tasks such as semantic segmentation.
    \item Automated visual attribute generation and feature selection techniques tailored for segmentation tasks.
\end{itemize}
\vspace{0.5em}
Our proposed \modelname framework is a plug-and-play approach that can improve the out-of-domain performance of existing OVSS models. We integrate our \modelname on multiple state-of-the-art OVSS approaches and evaluate across 22 segmentation datasets ranging across multiple domains (e.g., medical, agricultural, earth monitoring) and visual modalities (visible spectrum, electromagnetic, multi-spectral) establishing the state-of-the-art performance of our \modelname framework.

\section{Related Work}
\label{sec:related}

\begin{figure*}[t]
    \centering
    \includegraphics[width=1.0\linewidth]{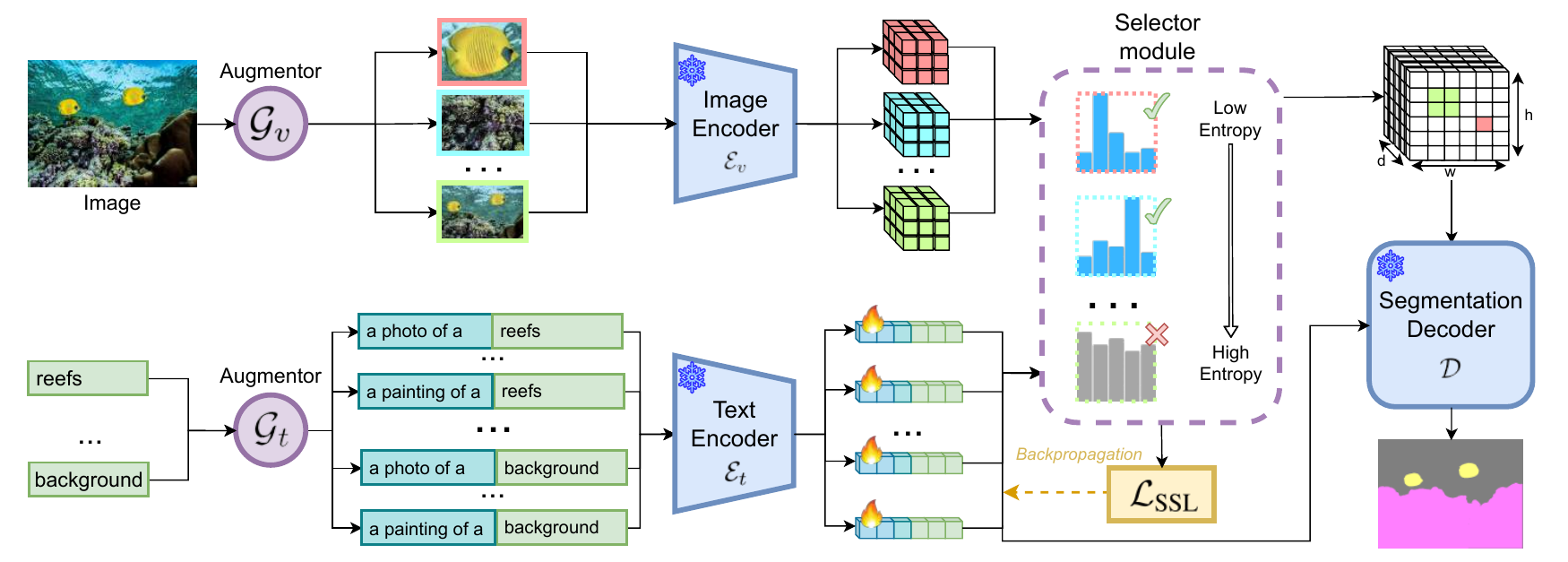}
    \vspace{-1.0em}
    \caption{
    \textbf{Overview of \modelname} (a) Our image embedding updating framework consists of filtering out confident image patches followed by updating the original image embedding. (b) Our test time optimization framework consists of updating prompts based on the most confident crops using backpropagation followed by the addition of attributes for generalization.}
    \label{fig:overall}
\end{figure*}

\noindent
\textbf{Zero-Shot Segmentation:} Contrastive vision language models \cite{clip,align} drive strong zero-shot performance in open-vocabulary semantic segmentation (OVSS) tasks \cite{catseg,ovseg,xu2023side} and empower models to learn segmentation from weak image-level supervision — eliminating the need for pixel-level human annotations \cite{clippy,clip-dinoiser,proxyclip}.  
However, performance of these approaches is limited to mainstream (in-domain) tasks, often suffering on specialized OVSS tasks \cite{MESS}. In fact, most approaches that generate competitive results in in-domain benchmarks \cite{catseg,ovseg,xu2022simple,xu2023side,zhang2023simple,zou2023generalized,ding2022decoupling} perform poorly in out-of-domain tasks when compared to their supervised counterparts \cite{MESS}. For example, best performing OVSS models achieve zero-shot accuracies almost 50\% below supervised counterparts on engineering, agriculture, or medical domain tasks \cite{bianchi2021corrosion,shivakumar2020pst900,bashkirova2022zerowaste,islam2020semantic,WahCUB_200_2011,haug2015crop}. 
Our proposed \modelname aims to bridge this gap using novel test-time optimization techniques and operates as a plug-and-play approach that improves the performance of both pixel-level and image-level supervised OVSS approaches on specialized domain tasks. To the best of our knowledge, \modelname is the first to explore test-time optimization in image segmentations settings adapting to specialized domains. 

\noindent
\textbf{Domain Adaptive Segmentation:}
Unsupervised domain adaptation for semantic segmentation approaches, particularly those focused on self-supervision and visual augmentation, is another line of closely related works \cite{Hoyer23,Hoyer22eccv,Hoyer22cvpr,Chen22,Lu22, Zhang21,Wang22,Kundu21, Li20, Zheng20, mata2024copt}. Contrastive losses to align representations together with augmentations-based view generations allow self-learning on unlabeled out-of-domain data. However, these approaches are limited to the visual modality performing segmentation on a closed set of fixed object categories that are known during training. In contrast, our \modelname framework can operate zero-shot on a range of open-vocabulary tasks.  

\noindent\textbf{Open-Vocabulary Domain Adaptation:}
Several recent works explore self-supervision or data augmentation for improved zero-shot performance of open vocabulary classification \cite{menon2022visual,llmmutate,jin2024llms,ozturk2024intelligent}. Textual attribute generation as a language modality augmentation improves model representation generality in \cite{menon2022visual,llmmutate}. Visual feature selection for domain adaptation is explored in \cite{ozturk2024intelligent}. However, these approaches are limited to classification settings and do not directly generalize to segmentation. 
OpenDAS \cite{yilmaz2024opendas} on the other hand focuses on open-vocabulary domain adaptation for segmentation but requires supervision to learn unlike ours. 
Contemporary work, PointSeg \cite{he2024pointseg}, performs test time optimization with projective geometry based adaptations for 3D segmentation tasks. 
In contrast, our proposed \modelname focuses on \textit{2D image space segmentation} specific adaptation using test-time optimization techniques. Closely related is TPT \cite{tpt} which optimizes a learnable prompt to adapt open-vocabulary classification models to various tasks. However, given the pixel-wise classification nature of segmentation and the presence of more than a single concept within an image that needs recognition (i.e., different pixels belonging to different categories need to be recognized), direct application of TPT \cite{tpt} to OVSS tasks is infeasible. Our \modelname explores unique pixel-level entropy calculations and multi-concept aware loss functions to perform test-time optimization for segmentation.   

\noindent\textbf{Language Modality Prompt Learning:}
Contrastive vision language models \cite{clip,align} exhibit strong sensitivity to prompt templates used for the language modality inputs during zero-shot probing \cite{clip}. Early prompt hand-crafting (in natural language) \cite{clip} was replaced by learnable prompt embeddings that learn task-specific prompts using labeled training data \cite{zhou2022coop, zhou2022cocoop}. The reliance on training data is eliminated in \cite{tpt} where prompt embeddings are optimized for each sample at test time using a self-supervised loss. This test-time prompt tuning is further improved for better generalization in \cite{abdul2024align,ma2024swapprompt,zhao2024testtime}. However, all of these approaches are primarily designed for classification tasks, as opposed to segmentation. Our proposed \modelname differs with its segmentation-specific test-time optimizations suited for adapting to specialized domain OVSS tasks.

\section{Methodology}
\label{sec:method}

In this section, we present our \modelname framework for specialized domain OVSS tasks. Given an existing model capable of OVSS, our goal is to adapt its representations to a specialized domain with only test-time optimization. In classification tasks, prompt tuning and feature selection techniques have proven effective for efficiently adjusting model representations, even at test time \cite{Jia2022VisualPT,zhou2022coop,tpt,ozturk2024intelligent,Shang2023ActiveVR}. Motivated by these, we propose test-time optimization (TTO) for jointly modifying both visual and textual features. We first construct a self-supervised loss suitable for measuring representation suitability for segmentation tasks. We then utilize this loss to modify visual representations while preserving their spatial structure which is crucial for segmentation. On the textual modality, we use our loss to guide gradient-based updates to modify per-category representations. We further augment category representations with visually relevant attributes pre-generated using a large language model (LLM). These attributes are filtered at test-time conditioned on the test sample. Finally, we send these domain-adapted representations to the OVSS model segmentation head to generate pixel-level predictions.  

In the following, we outline some background along with our architecture, describe our self-supervised objective, detail our modifications to representations on both modalities and finally present our overall \modelname framework that is a plug-and-play module over existing OVSS approaches.

\subsection{Background \& Architecture}
Given an image $\rmX \in \sR^{H \times W \times 3}$ and a set of category names $\sY=\{y_1, y_2, ... , y_n\}$, we aim to classify each of the $H \cdot W$ pixels of the image into one of the $n$ categories. In OVSS, the category set $\sY$ can be of arbitrary length and contains any category name defined in natural language. 

We define a generic pixel-level language-aligned representation learning OVSS model containing an image encoder ($\gE_v$), a text encoder ($\gE_t$), and a segmentation decoder ($\gD$). In general, the image encoder tends to be a CNN or ViT backbone while the text encoder is a transformer model. The segmentation decoder may vary across methods, with approaches such as \cite{clippy} using zero-shot probing at patch level similar to CLIP \cite{clip}, and others using specialized operations and learnable modules \cite{catseg}. Our framework aims to be agnostic to the segmentation decoder and focuses on modifying image and text encoder representations. 

In detail, we introduce a selector module that processes features from image and text encoders, calculates a self-supervised loss to guide the test-time feature optimization, and outputs domain-adapted features that can directly operate with the segmentation decoder. We additionally utilize two visual and textual \textit{augmentor} modules ($\gG_v$ and $\gG_t$) that allow extracting augmented versions of features from the encoders to feed to our selector module.
An overview of this architecture is presented in \Cref{fig:overall}.

\subsection{Test-Time Feature Optimization}
\label{subsection:ttfo}
The key role of our selector module is to modify representations to a form best suited to solving OVSS tasks in a given specialized domain. To this end, we propose a self-supervised loss that can guide such modifications. 

Consider a set of visual features $\sF_v = \{a_i \mid i \in [1, m]\} $  where $a_i = \gE_v (\tilde{\rmX}_i)$ and $\tilde{\rmX}_i$ are obtained by applying $m$ different visual augmentations onto the image $\rmX$. Note that each $\gE_v (\tilde{\rmX}_i) \in \sR^{h' \times w' \times d_v}$ where $h', w'$ are spatial dimensions and $d_v$ is the channel dimension.  
Also consider $p$ learnable prompts that are combined with each category $y_j$ to obtain $n$ (number of different categories) textual feature sets $\sF_{t,j} = \{b^j_k \mid k \in [1, p] \}$. Each feature $b^j_k \in \sR^{d_t}$ is from the textual encoder $\gE_t$. These features are also augmented using category attributes generated using a large language model (details in \Cref{subsec:text_attr}). 

We first define an entropy loss for each spatial location $q \in \sR^{h' \times w'}$ of each visual feature map $i$ as,
\begin{align}
    \gL_{\text{ent}}^{q, i} \left( \sF_v, \sF_{t,j} \right) =
    - \sum_{j=1}^{n} \sum_{k=1}^p \pr(b^j_k | a_i) \cdot \mathtt{log} \ \pr(b^j_k | a_i) 
\end{align}
and a cross entropy loss using pseudo-labels $\hat{y}$ (normalized cross-modal feature similarity) as, 
\begin{align}
    \gL_{\text{ce}}^{q, i} \left( \sF_v, \sF_{t,j} \right) =
    - \sum_{j=1}^{n} \sum_{k=1}^p \hat{y}[j] \cdot \mathtt{log} \ \pr(b^j_k \mid a_i)
\end{align}
where $\hat{y}[j]$ is its $j$th element. We also define P operator as,
\begin{align}
    \pr(b^j_k \mid a_i) = \frac{\mathtt{exp}(\mathtt{sim}(\mathit{b^j_k} \cdot \mathit{a_i})\tau)}{\sum_{j=1}^{K}\mathtt{exp}(\mathtt{sim}(\mathit{b^j_k} \cdot \mathit{a_i})\tau)} 
\end{align}
where $\tau$ is a temperature parameter and $\mathtt{sim}$ denotes a distance metric, which is cosine similarity in our implementation. We utilize the PCGrad operation ($\phi$) from \cite{yu2020gradient} to combine these two losses and obtain our complete self-supervised loss as in \Cref{eq:ssl}. The PCGrad operation reduces the effects of conflicting gradients in terms of their magnitude, direction and curvature by projecting the gradient of each task onto the normal plane of the gradient of the other task. This reduces the amount of opposing gradient interactions between the functions and ensures optimal gradient flow minimizing both loss functions during our test-time optimization. This leads to,
\begin{align}
    \gL_{\text{SSL}}^{q,i} &=  \phi \left(
        \gL_{\text{ent}}^{q, i} (\sF_v,\sF_{t,j}), \ \gL_{\text{ce}}^{q, i} ( \sF_v, \sF_{t,j} )
    \right) \label{eq:ssl1} \\
    \gL_{\text{SSL}}^q &= \gamma_{\text{sel}} \left( \{ \gL_{\text{SSL}}^{q, i} \mid i \in [1, m] \}
    \right) \label{eq:ssl2} \\
    \gL_{\text{SSL}} &= 
    \gamma_{\text{aggr}} \left( \{ \gL_{\text{SSL}}^{q} \mid q \in \sR^{h' \times w'} \} \right) 
\label{eq:ssl}
\end{align}
where $\gamma_{\text{sel}}$ performs visual feature selection and $\gamma_{\text{aggr}}$ operator performs spatial aggregation. 
Inputs to the loss functions ($\sF_v,\sF_{t,j}$) are omitted for clarity in \Cref{eq:ssl1,eq:ssl2,eq:ssl}. We hypothesize that higher $\gL_{\text{SSL}}$ values correspond to higher uncertainty and therein less informative features. Our intuition is that features minimizing $\gL_{\text{SSL}}$ would be the most informative set of features for a given task. 

In terms of the test-time optimization, we first describe the visual modality. The visual feature selection operation $\gamma_{\text{sel}}$ picks $m'$ good features. Entropy is spatially aggregated per feature (using mean operation following ablations) and the $m'$ least entropy features are selected as optimal. This follows our intuition for minimal $\gL_{\text{SSL}}$ corresponding to the most informative features. We resort to this selection as opposed to gradient-based updates given the need for retaining the spatial structure of features and the larger dimensionality of these features. We also perform re-scaling operations for aggregating the good features to ensure correct alignment across feature spatial dimensions (details in \Cref{subsec:vis_aggr}) which is necessary for the segmentation task. 

On the textual modality, each of our textual features $b^j_k$ (in $\sF_{t,j}$) is composed of two separate embeddings, $c_j$ and $g_k$, where $c_j$ is a category-specific embedding (for category $j$) and $g_k$ is a general category agnostic embedding (with $k$ different such general embeddings). Given our loss function in \Cref{eq:ssl}, we optimize these embeddings over $t$ iterations during test time to obtain textual features that are well-suited for each specialized domain. This optimization happens at a sample level, allowing the embeddings to adapt to each instance (i.e., image) being segmented. In contrast to classification approaches such as TPT \cite{tpt}, we utilize multiple category-specific learnable prompts. We hypothesize that learning such per-category prompts would better handle the multi-concept output nature of segmentation (i.e., to segment multiple categories within a single image). 

Having presented our test-time optimization strategy, we next discuss how LLM-generated category attributes are injected into our framework. 

\subsection{Category Attribute Aggregation}
\label{subsec:text_attr}
Visual attributes are the characteristics used to recognize and identify objects. For example, we identify an elephant by its large black body and long trunk. Similarly, such attributes can be leveraged to enhance OVSS performance in specialized domains where category names could be rare, obscure terminology (e.g. \texttt{mediastinum} in \Cref{fig:domain_analysis}). Modern LLMs, while limited to language modality, are known to contain knowledge regarding such obscure terms used across even some highly specialized domains \cite{Yu2023KoLACB}. 

For a given OVSS task, we feed the category names to such an LLM and generate sets of per-category attributes that are \textit{visually descriptive} of the object category and \textit{textually distinct} from other object categories. The latter is specifically important for segmentation in contrast to classification approaches. 
We explore a range of different LLMs as well as prompting styles (i.e., the same LLM would generate very different outputs for different styling of the same question) to generate an optimal set of category attributes. We also explore multiple \textit{templating} operations conditioned on category names for the generated attributes. Our experiments indicate that each of these hyper-parameters plays a significant role in how well the category attributes can contribute to overall performance improvements. We refer to \Cref{app:cat_attributes} for further details on attribute generation.   

\begin{table*}[t]
\centering
\small
\def\arraystretch{1.4}  
\setlength\tabcolsep{0.4em}  
\scalebox{0.69}{
\begin{tabular}{@{}lcccccc|ccccc|cccc|cccc|ccc|c@{}}
\toprule
 & \multicolumn{6}{c|}{General} & \multicolumn{5}{c|}{Earth Monitoring} & \multicolumn{4}{c|}{Medical Sciences} & \multicolumn{4}{c|}{Engineering} & \multicolumn{3}{c|}{Agri. and Biology} & \\
 & \rotatebox[origin=l]{90}{BDD100K \cite{yu2020bdd100k}} & \rotatebox[origin=l]{90}{Dark Zurich \cite{sakaridis2019guided}} & \rotatebox[origin=l]{90}{MHP v1 \cite{li2017multiple}} & \rotatebox[origin=l]{90}{FoodSeg103 \cite{wu2021large}} & \rotatebox[origin=l]{90}{ATLANTIS \cite{erfani2022atlantis}} & \rotatebox[origin=l]{90}{DRAM \cite{cohen2022semantic}} & \rotatebox[origin=l]{90}{iSAID \cite{waqas2019isaid}} & \rotatebox[origin=l]{90}{ISPRS Pots. \cite{alemohammad2020landcovernet}} & \rotatebox[origin=l]{90}{WorldFloods \cite{mateo2021towards}} & \rotatebox[origin=l]{90}{FloodNet \cite{rahnemoonfar2021floodnet}} & \rotatebox[origin=l]{90}{UAVid \cite{lyu2020uavid}} & \rotatebox[origin=l]{90}{Kvasir-Inst. \cite{jha2021kvasir}} & \rotatebox[origin=l]{90}{CHASE DB1 \cite{fraz2012ensemble}} & \rotatebox[origin=l]{90}{CryoNuSeg \cite{mahbod2021cryonuseg}} & \rotatebox[origin=l]{90}{PAXRay-4 \cite{seibold2022detailed}} & \rotatebox[origin=l]{90}{Corrosion CS \cite{bianchi2021corrosion}} & \rotatebox[origin=l]{90}{DeepCrack \cite{liu2019deepcrack}} & \rotatebox[origin=l]{90}{PST900 \cite{shivakumar2020pst900}} & \rotatebox[origin=l]{90}{ZeroWaste-f \cite{bashkirova2022zerowaste}} & \rotatebox[origin=l]{90}{SUIM \cite{islam2020semantic}} & \rotatebox[origin=l]{90}{CUB-200 \cite{WahCUB_200_2011}} & \rotatebox[origin=l]{90}{CWFID \cite{haug2015crop}} & \rotatebox[origin=l]{90}{Mean} \\
\midrule
\textit{Random} & \textit{1.48} & \textit{1.31} & \textit{1.27} & \textit{0.23} & \textit{0.56} & \textit{2.16} & \textit{0.56} & \textit{8.02} & \textit{18.43} & \textit{3.39} & \textit{5.18} & \textit{27.99} & \textit{27.25} & \textit{31.25} & \textit{31.53} & \textit{9.3} & \textit{26.52} & \textit{4.52} & \textit{6.49} & \textit{5.3} & \textit{0.06} & \textit{13.08} & \textit{10.27} \\
\textit{Best sup.} & \textit{44.8} & \textit{63.9} & \textit{50.0} & \textit{45.1} & \textit{42.22} & \textit{45.71} & \textit{65.3} & \textit{87.56} & \textit{92.71} & \textit{82.22} & \textit{67.8} & \textit{93.7} & \textit{97.05} & \textit{73.45} & \textit{93.77} & \textit{49.92} & \textit{85.9} & \textit{82.3} & \textit{52.5} & \textit{74.0} & \textit{84.6} & \textit{87.23} & \textit{70.99} \\
\hline
\hline
ZSSeg-B \cite{xu2022simple} & 32.36 & 16.86 & 7.08 & 8.17 & 22.19 & 33.19 & 3.80 & 11.57 & 23.25 & 20.98 & 30.27 & 46.93 & 37.00 & 38.70 & 44.66 & 3.06 & 25.39 & 18.76 & 8.78 & 30.16 & 4.35 & 32.46 & 22.73 \\
ZegFormer-B \cite{ding2022decoupling} & 14.14 & 4.52 & 4.33 & 10.01 & 18.98 & 29.45 & 2.68 & 14.04 & 25.93 & 22.74 & 20.84 & 27.39 & 12.47 & 11.94 & 18.09 & 4.78 & 29.77 & 19.63 & 17.52 & 28.28 & 16.8 & 32.26 & 17.57 \\
X-Decoder-T \cite{zou2023generalized} & 47.29 & 24.16 & 3.54 & 2.61 & 27.51 & 26.95 & 2.43 & 31.47 & 26.23 & 8.83 & 25.65 & 55.77 & 10.16 & 11.94 & 15.23 & 1.72 & 24.65 & 19.44 & 15.44 & 24.75 & 0.51 & 29.25 & 19.80 \\
SAN-B \cite{xu2023side} & 37.40 & 24.35 & 8.87 & 19.27 & 36.51 & 49.68 & 4.77 & 37.56 & 31.75 & 37.44 & 41.65 & 69.88 & 17.85 & 11.95 & 19.73 & 3.13 & 50.27 & 19.67 & 21.27 & 22.64 & 16.91 & 5.67 & 26.74 \\
OpenSeeD-T \cite{zhang2023simple} & 47.95 & 28.13 & 2.06 & 9.00 & 18.55 & 29.23 & 1.45 & 31.07 & 30.11 & 23.14 & 39.78 & 59.69 & 46.68 & 33.76 & 37.64 & 13.38 & 47.84 & 2.50 & 2.28 & 19.45 & 0.13 & 11.47 & 24.33 \\
Gr.-SAM-B \cite{ren2024grounded} & 41.58 & 20.91 & 29.38 & 10.48 & 17.33 & 57.38 & 12.22 & 26.68 & 33.41 & 19.19 & 38.34 & 46.82 & 23.56 & 38.06 & 41.07 & 20.88 & 59.02 & 21.39 & 16.74 & 14.13 & 0.43 & 38.41 & 28.52 \\
CAT-Seg-B \cite{catseg} & 44.58 & 27.36 & 20.79 & 21.54 & 33.08 & 62.42 & 15.75 & 41.89 & 39.47 & 35.12 & 40.62 & 70.68 & 25.38 & 25.63 & 44.94 & 13.76 & 49.14 & 21.32 & 20.83 & 39.10 & 3.40 & 45.47 & 33.74 \\  
\rowcolor{Gray}
CAT-Seg-B-TTO & 44.03 & 27.97 & 21.37 & 22.48 & 33.50 & 65.12 & 18.59 & 42.56 & 39.97 & 36.83 & 40.89 & 70.85 & 32.33 & 33.41 & 45.98 & 21.56 & 53.52 & 21.58 & 20.85 & 39.86 & 3.40 & 45.72 & 35.56\inc{1.8} \\ \hline
OVSeg-L \cite{ovseg} & 45.28 & 22.53 & 6.24 & 16.43 & 33.44 & 53.33 & 8.28 & 31.03 & 31.48 & 35.59 & 38.8 & 71.13 & 20.95 & 13.45 & 22.06 & 6.82 & 16.22 & 21.89 & 11.71 & 38.17 & 14.00 & 33.76 & 26.94 \\
SAN-L \cite{xu2023side} & 43.81 & 30.39 & 9.34 & 24.46 & 40.66 & 68.44 & 11.77 & 51.45 & 48.24 & 39.26 & 43.41 & 72.18 & 7.64 & 11.94 & 29.33 & 6.83 & 23.65 & 19.01 & 18.32 & 40.01 & 19.30 & 1.91 & 30.06 \\
Gr.-SAM-L \cite{ren2024grounded} & 42.69 & 21.92 & 28.11 & 10.76 & 17.63 & 60.80 & 12.38 & 27.76 & 33.40 & 19.28 & 39.37 & 47.32 & 25.16 & 38.06 & 44.22 & 20.88 & 58.21 & 21.23 & 16.67 & 14.30 & 0.43 & 38.47 & 29.05 \\ 
CAT-Seg-L \cite{catseg} & 45.83 & 33.10 & 30.03 & 30.47 & 33.60 & 66.54 & 16.09 & 51.42 & 49.86 & 39.84 & 42.02 & 68.10 & 24.99 & 35.06 & 54.50 & 16.87 & 31.42 & 25.26 & 30.62 & 53.94 & 9.24 & 39.00 & 37.63 \\ 
\rowcolor{Gray}
CAT-Seg-L-TTO & 46.78 & 34.58 & 32.27 & 31.16 & 34.07 & 70.24 & 19.81 & 52.55 & 49.15 & 39.79 & 42.41 & 74.05 & 29.96 & 42.90 & 58.69 & 21.40 & 32.27 & 25.86 & 32.80 & 57.77 & 9.97 & 47.47 & 40.27\inc{2.6} \\
\bottomrule
\end{tabular}
}
%
\caption{
\textbf{Zero-Shot Semantic Segmentation on Out-of-Domain Datasets:}
Our proposed \modelname achieves state-of-the-art performance across 22 different datasets on the MESS benchmark highlighting its strong generality across domains. 
}
\label{tab:tbl_main}
\vspace{0.5em}
\end{table*}


\begin{table*}[t]
\centering
\small
\def\arraystretch{1.4}  
\setlength\tabcolsep{1.2em}  
\scalebox{0.85}{

\begin{tabular}{llllll|l}
\toprule
 & General & Earth Monitoring & Medical Sciences & Engineering & Agri. \& Biology & Mean\\
\midrule
\textit{Random} & \textit{1.17} & \textit{7.12} & \textit{29.51} & \textit{11.71} & \textit{6.51} & \textit{10.27} \\
\textit{Best sup.} & \textit{48.62} & \textit{79.12} & \textit{89.49} & \textit{67.66} & \textit{81.94} & \textit{70.99} \\
\hline
CLIPpy \cite{clippy} & 10.79 & 19.62 & 30.39 & 10.10 & 19.27 & 17.39\\ 
CLIP-DINOiser \cite{clip-dinoiser} 
                   & 25.77         & 26.87          & 42.65          & 33.74          & 30.15          & 31.14 \\ \rowcolor{Gray}
CLIP-DINOiser-TTO & 26.17\inc{0.4} & 27.94\inc{1.1} & 48.02\inc{5.4} & 34.76\inc{1.0} & 30.84\inc{0.7} & 32.74\inc{1.6} \\
\bottomrule
\end{tabular}
}
%
\caption{
\textbf{Zero-Shot Unsupervised Semantic Segmentation on Out-of-Domain Datasets:}
We evaluate mask-free training methods and a variant of our \modelname trained under similar settings. These approaches utilize no pixel-level human annotations and only image-level captions from noisy internet-scale datasets (same data used to train CLIP \cite{clip}). Our proposed \modelname achieves state-of-the-art performance under these settings as well. 
}
\label{tab:uss}
\end{table*}

Given a set of generated per-category attributes, $\sA_j = \{u^j_r \mid r \in [1, s_j] \}$, we first apply an attribute feature aggregation operation to emphasize more relevant attributes. First, we take the cosine similarity between each attribute's normalized text embedding $\hat{\gE_t}({u^j_r})$ and corresponding normalized category name ($y_j$) learned embedding $\hat{b_j}$, as $\gamma_{\text{cs}}(u^j_r, y_j)$ where $\hat{\gE_t}$ denotes channel-dimension normalization of text encoder outputs. We weight each attribute by this cosine similarity to reflect how closely the attribute is related to the class, ensuring that more relevant attributes contribute more significantly to the final attribute embedding and calculate an averaged embedding as follows, 
\begin{equation}
    \gamma_{\text{attr}}(\sA_j) = \frac{\sum_{r=1}^{s_j} \gamma_{\text{cs}}(u^j_{r},y_j) \cdot \hat{\gE_t}({u^j_{r}})}{\left\|\sum_{r=1}^{s_j} \gamma_{\text{cs}}(u^j_{r},y_j) \cdot \hat{\gE_t}({u^j_{r}})\right\|}
\end{equation}
where $\gamma_{\text{attr}}(\sA_j) \in \sR^{d_t}$ is our aggregated attribute-aware embedding for category $j$. 
To obtain the final text embedding for a given image $\rmX$, we calculate a weighted average of our tuned text embeddings $\{b^j_k \mid k \in [1, p] \}$ for category $j$ (see \Cref{subsection:ttfo}) with our aggregated attribute-aware embedding $\gamma_{\text{attr}}(\sA_j)$ as, 
\begin{align}
    \rvf_t^j = \frac{\beta}{p} \sum_{k=1}^p b^j_k + (1 - \beta) \gamma_{\text{attr}}(\sA_j)
\label{eq:final_txt_embd}
\end{align}
where $\beta$ is a hyper-parameter which we fix experimentally and $\rvf_t^j$ is our final text embedding for category $j$. We obtain embeddings for all $n$ categories as $\rmF_t = [\rvf_t^1, \rvf_t^2, ... , \rvf_t^n]$ the final text embeddings for probing the given image $\rmX$.

\subsection{Visual Feature Aggregation}
\label{subsec:vis_aggr}

Let $a_{orig}$ be the original image embedding. We interpolate spatial dimensions of $a_{orig}$ to the original image size and filtered $m'$ image embeddings to their post-augmentation sizes. We then update $a_{orig}$ using $\{a_i \mid i \in [1,m']\}$. 
\begin{align}
    a'_{orig} =
    \sum_{j'=h^i_1}^{h^i_2} \sum_{k'=w^i_1}^{w^i_2} a_{orig}^{j',k'} + a_i^{j'-h^i_1,k'-w^i_1}
\label{eq:vfa}
\end{align}
for $(h^i_1,w^i_1,h^i_2,w^i_2)$ bounding coordinates of $a_i$ when aligned to the original image location (e.g., when augmentation involves a crop of an image subregion). Similarly, we aggregate all $m'$ to obtain the aggregated visual feature $a'_{orig}$. Next, we obtain our final visual embedding $\rvf_v$ as,
\begin{table*}[t]
\centering

\begin{minipage}{0.44\textwidth}  
\centering
\small
\def\arraystretch{1.0}  
\setlength\tabcolsep{0.9em}  
\scalebox{0.78}{  
\begin{tabular}{l|cccccc}
\toprule
Method   &  CAA   & VFA   & TTO   & ZWF  & DZ & DRAM   \\ \midrule
CAT-Seg-L & \xmark & \xmark & \xmark & 30.6 & 33.1 & 66.5 \\
Ours     & \cmark & \xmark & \xmark & 31.8 & 33.5 & 69.2 \\
Ours     & \cmark & \cmark & \xmark & 31.9 & 34.2 & 69.8\\ \rowcolor{Gray}
Ours     & \cmark & \cmark & \cmark & 32.8 & 34.6 & 70.2\\ \bottomrule
\end{tabular}
}
\caption{\footnotesize
\textbf{Framework Ablation}: 
We ablate each component of \modelname: category attribute aggregation (CAA), visual feature aggregation (VFA), and test-time optimization (TTO). We report mIoU (\%) on ZeroWaste-F (ZWF) \cite{bucher2019zero}, Dark Zurich (DZ) \cite{sakaridis2019guided} and DRAM \cite{cohen2022semantic} datasets highlighting the individual contribution of each component. 
}
\label{tbl:main_ablation}
\end{minipage}
%
\hspace{0.01\textwidth}
\begin{minipage}{0.54\textwidth}  
\centering
\small
\def\arraystretch{1.2}  
\setlength\tabcolsep{0.5em}  
\scalebox{0.80}{
\begin{tabular}{l|l|l|l|l|l|l}
\toprule
Method      & BDD  & DRAM & ZWF   & KI   & DC   & Avg (all 22)  \\ \midrule
CAT-Seg      & 45.8 & 66.5 & 30.6 & 68.1 & 31.4 & 38.1 \\
CAT-Seg+Prompts  & 27.6\dec{18.2} & 34.9\dec{31.6} & 18.2\dec{12.4} & 50.6\dec{17.5} & 29.9\dec{1.5} & 30.9\dec{7.2} \\ \rowcolor{Gray}
\modelname (ours)        & 46.8\inc{1.0} & 70.2\inc{3.7} & 32.8\inc{2.2} & 74.1\inc{6.0} & 32.3\inc{0.9} & 40.2\inc{2.1} \\ \bottomrule
\end{tabular}
}
\caption{\footnotesize
\textbf{Prompt Ablation}:
We explore naively injecting prompts into the CAT-Seg baseline (row 2) without our TTO component. Such naive prompt injection does not lead to improvements similar to our \modelname. In fact, it reduces performance as the model has not been trained to operate with such prompts. We particularly highlight datasets where large performance drops occur while \modelname shows improvement. 
}
\label{ablate:prompts}
\end{minipage}


\vspace{1.0em}


\begin{minipage}{0.32\textwidth}
    \centering
    \small
    \def\arraystretch{1.1}  
    \setlength\tabcolsep{0.9em}  
    \scalebox{0.80}{
    \begin{tabular}{l|l|l}
    \toprule
    Method               & DZ             &  CHASE           \\ \midrule
    CAT-Seg               & 33.1           &  25.0            \\
    + Naive-TTO \cite{tpt} & 32.8\dec{0.3}  &  24.2\dec{0.8}   \\ \rowcolor{Gray}
    \modelname (ours)    & 34.6\inc{1.5}  &  30.0\inc{5.0}   \\ \bottomrule
    \end{tabular}}
    \caption{\footnotesize
    \textbf{TTO Ablation:}
    TTO techniques for classification \cite{tpt} do not work well for segmentation. \modelname achieves improvements through multiple segmentation specific design choices.}
    \label{ablate:tto}
    \vspace{-1em}
\end{minipage}
%
\hspace{0.01\textwidth}
\begin{minipage}{0.32\textwidth}
    \centering
    \small
    \def\arraystretch{1.0}  
    \setlength\tabcolsep{0.7em}  
    \scalebox{0.80}{
    \begin{tabular}{l|l|l}
    \toprule
    Method                 & DZ            & KI            \\ \midrule
    CAT-Seg                & 33.1          & 68.1          \\
    Only TTO+VFA (ours)         & 34.1\inc{1.0} & 71.0\inc{2.9} \\ 
    Only CAA (ours)        & 33.5\inc{0.4} & 70.4\inc{2.3} \\ \rowcolor{Gray}
    \modelname (ours-full) & 34.6\inc{1.5} & 74.1\inc{6.0} \\ \bottomrule
    \end{tabular}}
    \caption{\footnotesize
    \textbf{Textual vs Visual:}
    Both only CAA (aggregating LLM-generated prompts) and only TTO+VFO improve performance, but their joint application leads to even further gains.
    }
    \label{ablate:tto_prompt}
    \vspace{-1em}
\end{minipage}
%
\hspace{0.01\textwidth}
\begin{minipage}{0.15\textwidth}  
\centering
\small
\def\arraystretch{1.0}  
\setlength\tabcolsep{0.7em}  
\scalebox{0.80}{  
\begin{tabular}{l|c}
\toprule
Aggregation & DZ  \\ \midrule
Attr. tuning       & 33.3 \\
Post-Aggr.         & 34.3 \\  \rowcolor{Gray}
Pre-Aggr.          & 34.6 \\  
\bottomrule
\end{tabular}
}
\caption{\footnotesize
Attribute pre-aggregation leads to optimal performance. 
}
\label{ablate:attr_aggregation}
\end{minipage}
%
\hspace{0.01\textwidth}
\begin{minipage}{0.15\textwidth}
\centering
\small
\def\arraystretch{1.0}  
\setlength\tabcolsep{0.9em}  
\scalebox{0.80}{
\begin{tabular}{l|c}
\toprule
Tune   & DZ   \\ \midrule
 PE    &  32.8 \\ 
 CE    &  32.7 \\ \rowcolor{Gray}
 PE+CE &  34.6 \\ \bottomrule
\end{tabular}
}
\caption{\footnotesize
Our joint embedding tuning (row 3) gives top results.
}
\label{ablate:tpt_learneable_component}
\end{minipage}

\vspace{1.0em}
\end{table*}
\begin{align}
    \rvf_v = \gN(a_{orig})
\end{align}
where $\gN$ stands for normalization based on the number of times each pixel was updated and interpolating the image embedding back to the original spatial dimension of $a_{orig}$ (more details in \Cref{app:vis_aggr}). This process retains the spatial structure of the visual feature map while enhancing the objects present in the image.
This exact overall operation is used as $\gamma_{\text{aggr}}$ in \Cref{eq:ssl}.

Having obtained domain-adapted visual and textual features ($\rvf_v$ and $\rvf_t$ respectively), we calculate the final image segmentation as, 
\begin{align}
    \rmY = \gD(\rvf_v, \rmF_t)
\end{align}
where $\rmY$ corresponds to a segmentation for image $\rmX$ and $\gD$ is the segmentation decoder. 

\section{Experiments}
\label{sec:experiments}
In this section, we first describe our experimental setup and implementation details. Then we present evaluations across 22 specialized domain datasets from MESS benchmark \cite{MESS}
comparing against prior work to establish the state-of-the-art performance of our \modelname framework. Finally, we discuss our ablative studies highlighting the contributions of each design decision in our implementation. We discuss these in detail in \Cref{app:dataset}.   

\noindent \textbf{Implementation Details:} 
Our framework uses $p=5$ for number of prompts, $m=64$ for number of visual augmentations, and $m'$ as a variable such that the lowest 20\% entropy among the $m$ visual views is retained. 
We apply \modelname over baselines from CAT-Seg \cite{catseg} and CLIP-DINOiser \cite{clip-dinoiser}. For each setting, we utilize the relevant image and text encoders as well as segmentation decoders from the baseline.
For the optimization process, we employ separate step counts of 2 and 3 for entropy and cross-entropy losses respectively using PCGrad \cite{yu2020gradient} for joint updates. We use an AdamW optimizer with a learning rate of 5e-3. We tune hyperparameters using two held-out datasets and evaluate across all datasets and model variants using the same, fixed hyperparameters. We use two 24GB NVIDIA RTX A5000 or 16GB NVIDIA Quadro RTX 5000 GPUs for all experiments. Inference per image takes 1.5 seconds for \modelname (vs 0.5 seconds for CAT-Seg). In an open-vocabulary setting, increasing performance freely (unsupervised) is a challenging task and we achieve up to 27\% improvements (7.0\% on average) with this inference cost. 

\subsection{Semantic Segmentation}
CAT-Seg \cite{catseg} is a state-of-the-art open-vocabulary segmentation model trained with pixel-level annotations. We integrate our \modelname framework with both base and large variants of CAT-Seg \cite{catseg} and report these results in \Cref{tab:tbl_main}. Our approach consistently improves performance across both variants, with our large variant setting establishing a new state-of-the-art on the MESS benchmark.

\begin{figure*}[t]
    \centering
    \includegraphics[width=1\textwidth]{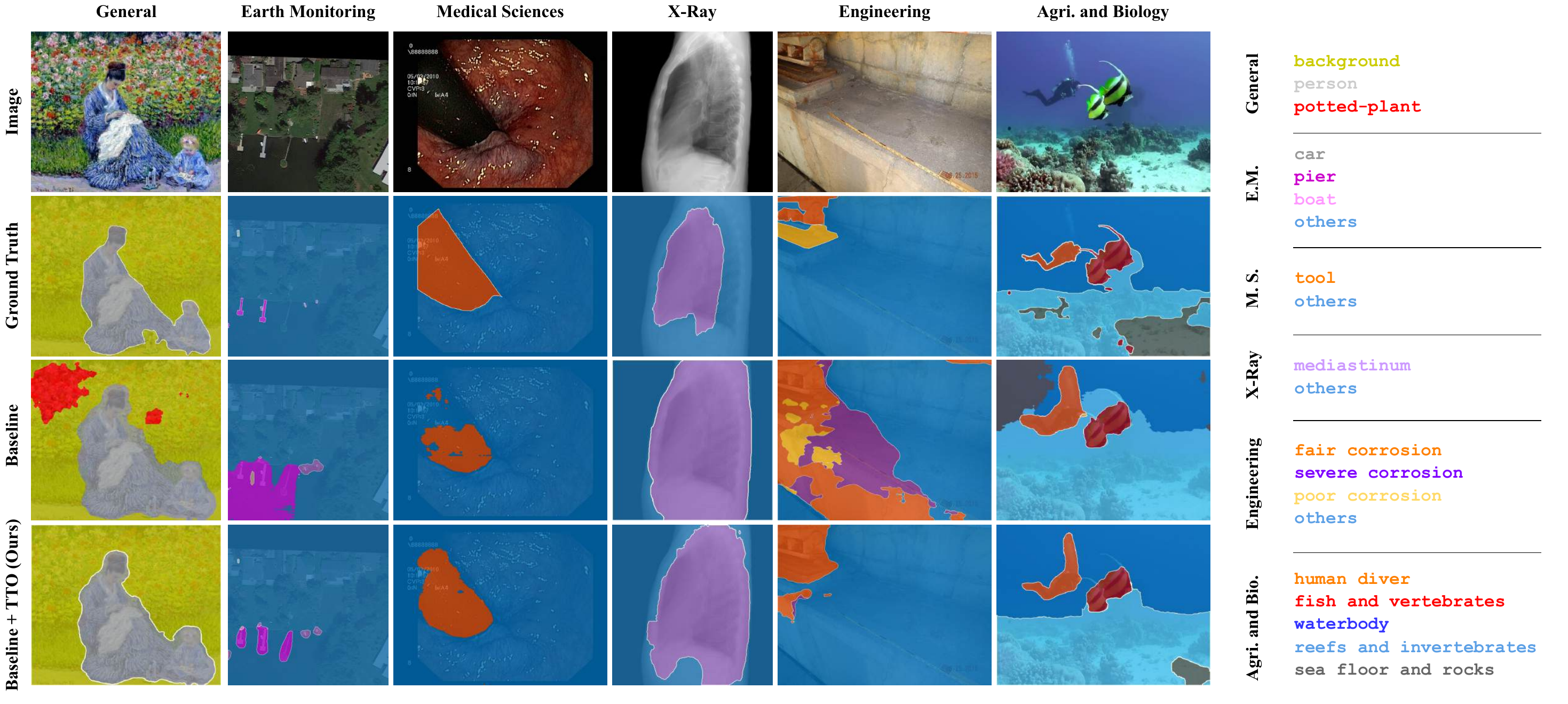}
    \caption{\textbf{Qualitative Evaluation:}
    Our proposed \modelname outperforms state-of-the-art CAT-Seg \cite{catseg} across diverse specialized-domain OVSS tasks as illustrated. We highlight the highly technical nature of some specialized domain category names (e.g., \texttt{mediastinum} under X-Ray). Our category attributes allow models to better understand such objects.  
    }
    \label{fig:domain_analysis}
\end{figure*}

Open-vocabulary segmentation in niche domains—such as those represented in the MESS benchmark (see \Cref{fig:domain_analysis} for sample images)—remains a challenging task. State-of-the-art segmentation methods achieve below 40 mIoU on these benchmarks \cite{catseg,ovseg}. Given this difficulty, even modest improvements of 1-2 mIoU are highly significant. Our \modelname framework demonstrates gains across 22 datasets, with improvements exceeding 27\% over baseline on certain datasets. 
Particularly with the stronger large variant, \modelname achieves \textit{clear and consistent improvements} with a 2.6 mIoU increase.
To put this into context, previous works such as SAN-L \cite{xu2023side} and Gr.SAM-L \cite{ren2024grounded} differ by only 1 mIoU, as seen in \Cref{tab:tbl_main}. 
These results underscore the effectiveness of \modelname in improving segmentation performance in challenging, zero-shot settings.

\subsection{Unsupervised Semantic Segmentation}
We next explore unsupervised semantic segmentation (training without pixel-wise annotations) within specialized domain tasks, a setting that has not been extensively studied. To the best of our knowledge, we are the first to explore this task.
We first evaluate two state-of-the-art unsupervised methods, CLIPpy \cite{clippy} and CLIP-DINOiser \cite{clip-dinoiser}, on the MESS benchmark as our baselines. We then integrate \modelname with CLIP-DINOiser and report results in \Cref{tab:uss}. Our framework achieves performance improvements in all domains, with a 1.6 increase in average mIoU.

Given the challenging nature of both unsupervised segmentation and specialized domain tasks, these gains are particularly noteworthy. Importantly, \modelname relies on no extended training time and \textit{no additional training data}. Instead, it employs a test-time optimization process using only the inputs available at inference. This data efficiency further highlights the significance of our results, demonstrating that \modelname is an effective strategy for improving segmentation performance under unsupervised settings.

\subsection{Ablative Study}
We now present extensive ablations of our proposed \modelname framework to establish its effectiveness and highlight the significance of our various design choices. 

\noindent \textbf{Framework Ablation}:
Our \modelname is composed of 3 individual components: category attribute aggregation (CAA), visual feature aggregation (VFA), and test-time optimization (TTO). We ablate these in \Cref{tbl:main_ablation}. Our results are consistent across three different datasets, highlighting each component's clear effectiveness.
In the case of CAA, we hypothesize that attributes assist in identifying rare classes as well as visually novel instances of general classes. However,  we note the importance of attribute quality for performance: particularly detail and content to differentiate from other classes are important. We provide more details on the importance of quality attributes in \Cref{app:attr_quality}. 
We hypothesize that VFA assists in isolating objects from the background similar to how it helps us to identify exact object boundaries when we zoom into an image. 
The purpose of TTO is to align the embeddings to the objects of interest in the image at hand. We take our ablation results as an indication of the successful contribution of these components to our overall \modelname framework.

\noindent \textbf{Prompt Ablation}:
Our proposed \modelname framework utilizes category attribute descriptions generated from an LLM to augment prompts used with open-vocabulary models. We investigate if these augmented prompts alone can help strengthen a baseline and whether modifications are necessary for these augmented prompts to be effective. Results presented in \Cref{ablate:prompts} indicate how naive injection of augmented prompts in fact hurts performance of baselines while our \modelname leads to consistent improvements. We hypothesize that existing models are not trained to handle such highly descriptive augmented prompts, leading to reduced performance. On the other hand, our feature aggregation and test-time optimization processes in \modelname allows models to adapt to handling such prompts much better, leading to improved performance. 

\noindent \textbf{TTO Ablation}:
As described in \Cref{sec:method}, our \modelname framework is designed specifically for segmentation with suitable embedding aggregation and optimization objectives. We compare these design choices against a state-of-the-art test-time optimization techniques for classification (TPT \cite{tpt}). We experiment by providing the same prompts with only the test-time loss calculation being replaced with \cite{tpt}. We report these results in \Cref{ablate:tto}. Our approach shows clear improvements while applying TPT \cite{tpt} naively on segmentation tasks leads to performance drops. We attribute this weaker performance to key differences in segmentation (needs spatially awareness and contains multiple concepts in a single image) that the TPT \cite{tpt} algorithm is not designed to handle. In contrast, our segmentation specific design choices lead to strong performance improvements over the baseline.

\noindent \textbf{Textual vs Visual Ablation}:
Our \modelname framework contains visual modality focused visual feature aggregation and test-time optimization as well as textual modality focused category attribute aggregation (CAA; uses category attributes from an LLM that are generated one time and stored). We explore how each sub-group performs independently and report these results in \Cref{ablate:tto_prompt}. Results indicate how each sub-group leads to performance improvements, while their joint application leads to additional gains. We also highlight how \modelname can operate without its CAA module (i.e. no LLM augmented prompts) to boost segmentation performance.

\noindent \textbf{Additional Ablations}:
We also present ablations on our design choices in \Cref{ablate:attr_aggregation,ablate:tpt_learneable_component}.  Results indicate that design choices in our \modelname framework lead to optimal performance in contrast to other common methods. We refer the reader to \Cref{app:additional_ablations} for more information.

\section{Conclusion}
\label{sec:conclusion}

In this work, we introduced \modelname, a novel test-time optimization framework to enhance open-vocabulary semantic segmentation (OVSS), particularly in highly-specialized domains. We address challenges of domain shifts in both visual and textual modalities by leveraging self-supervised objectives, LLM augmented textual attributes, learnable text embeddings, and locality-preserving feature aggregation techniques. By aligning model parameters with input images conditioned on task categories at test time, \modelname significantly improves segmentation accuracy in zero-shot settings without additional training data.
Extensive evaluation across 22 challenging OVSS datasets demonstrates the effectiveness of \modelname, with consistent improvements across diverse domains such as medical imaging, agriculture, and earth monitoring. 
The results establish \modelname as the first test-time optimization framework for OVSS, providing a plug-and-play solution that improves out-of-domain generalization for existing segmentation models. 

The limitation of \modelname is slower inference speed. We hope to explore distillation into lightweight models for faster inference as a future direction.
We hope our \modelname inspires future research in test-time optimization and its applications in real-world segmentation challenges.



{
    \small
    \bibliographystyle{ieeenat_fullname}
    \bibliography{main}

\begin{thebibliography}{69}
\providecommand{\natexlab}[1]{#1}
\providecommand{\url}[1]{\texttt{#1}}
\expandafter\ifx\csname urlstyle\endcsname\relax
  \providecommand{\doi}[1]{doi: #1}\else
  \providecommand{\doi}{doi: \begingroup \urlstyle{rm}\Url}\fi

\bibitem[Abdul~Samadh et~al.(2024)Abdul~Samadh, Gani, Hussein, Khattak, Naseer, Shahbaz~Khan, and Khan]{abdul2024align}
Jameel Abdul~Samadh, Mohammad~Hanan Gani, Noor Hussein, Muhammad~Uzair Khattak, Muhammad~Muzammal Naseer, Fahad Shahbaz~Khan, and Salman~H Khan.
\newblock Align your prompts: Test-time prompting with distribution alignment for zero-shot generalization.
\newblock \emph{NeurIPS}, 36, 2024.

\bibitem[AI@Meta(2024)]{llama3modelcard}
AI@Meta.
\newblock Llama 3 model card.
\newblock 2024.

\bibitem[Alemohammad and Booth(2020)]{alemohammad2020landcovernet}
Hamed Alemohammad and Kevin Booth.
\newblock Landcovernet: A global benchmark land cover classification training dataset.
\newblock \emph{arXiv preprint arXiv:2012.03111}, 2020.

\bibitem[Bashkirova et~al.(2022)Bashkirova, Abdelfattah, Zhu, Akl, Alladkani, Hu, Ablavsky, Calli, Bargal, and Saenko]{bashkirova2022zerowaste}
Dina Bashkirova, Mohamed Abdelfattah, Ziliang Zhu, James Akl, Fadi Alladkani, Ping Hu, Vitaly Ablavsky, Berk Calli, Sarah~Adel Bargal, and Kate Saenko.
\newblock Zerowaste dataset: Towards deformable object segmentation in cluttered scenes.
\newblock In \emph{CVPR}, pages 21147--21157, 2022.

\bibitem[Bianchi and Hebdon(2021)]{bianchi2021corrosion}
Eric Bianchi and Matthew Hebdon.
\newblock Corrosion condition state semantic segmentation dataset.
\newblock \emph{University Libraries, Virginia Tech: Blacksburg, VA, USA}, 3, 2021.

\bibitem[Blumenstiel et~al.(2023)Blumenstiel, Jakubik, Kuhne, and Vossing]{MESS}
Benedikt Blumenstiel, Johannes Jakubik, Hilde Kuhne, and Michael Vossing.
\newblock What a mess: Multi-domain evaluation of zero-shot semantic segmentation.
\newblock \emph{ArXiv}, abs/2306.15521, 2023.

\bibitem[Bucher et~al.(2019)Bucher, Vu, Cord, and P{\'e}rez]{bucher2019zero}
Maxime Bucher, Tuan-Hung Vu, Matthieu Cord, and Patrick P{\'e}rez.
\newblock Zero-shot semantic segmentation.
\newblock \emph{NeurIPS}, 32, 2019.

\bibitem[Chen et~al.(2022)Chen, Rong, Guo, Han, Sun, Xu, and Huang]{Chen22}
Runfa Chen, Yu Rong, Shangmin Guo, Jiaqi Han, Fuchun Sun, Tingyang Xu, and Wenbing Huang.
\newblock Smoothing matters: Momentum transformer for domain adaptive semantic segmentation, 2022.

\bibitem[Chiquier et~al.(2024)Chiquier, Mall, and Vondrick]{llmmutate}
Mia Chiquier, Utkarsh Mall, and Carl Vondrick.
\newblock Evolving interpretable visual classifiers with large language models.
\newblock \emph{arXiv preprint arXiv:2404.09941}, 2024.

\bibitem[Cho et~al.(2024)Cho, Shin, Hong, An, Lee, Arnab, Seo, and Kim]{catseg}
Seokju Cho, Heeseong Shin, Sung‐Jin Hong, Seungjun An, Seungjun Lee, Anurag Arnab, Paul~Hongsuck Seo, and Seung~Wook Kim.
\newblock Cat-seg: Cost aggregation for open-vocabulary semantic segmentation.
\newblock In \emph{CVPR}, 2024.

\bibitem[Cohen et~al.(2022)Cohen, Newman, and Shamir]{cohen2022semantic}
Nadav Cohen, Yael Newman, and Ariel Shamir.
\newblock Semantic segmentation in art paintings.
\newblock In \emph{Computer graphics forum}, pages 261--275. Wiley Online Library, 2022.

\bibitem[Ding et~al.(2022)Ding, Xue, Xia, and Dai]{ding2022decoupling}
Jian Ding, Nan Xue, Gui-Song Xia, and Dengxin Dai.
\newblock Decoupling zero-shot semantic segmentation.
\newblock In \emph{CVPR}, pages 11583--11592, 2022.

\bibitem[Erfani et~al.(2022)Erfani, Wu, Wu, Wang, and Goharian]{erfani2022atlantis}
Seyed Mohammad~Hassan Erfani, Zhenyao Wu, Xinyi Wu, Song Wang, and Erfan Goharian.
\newblock Atlantis: A benchmark for semantic segmentation of waterbody images.
\newblock \emph{Environmental Modelling \& Software}, 149:\penalty0 105333, 2022.

\bibitem[Fraz et~al.(2012)Fraz, Remagnino, Hoppe, Uyyanonvara, Rudnicka, Owen, and Barman]{fraz2012ensemble}
Muhammad~Moazam Fraz, Paolo Remagnino, Andreas Hoppe, Bunyarit Uyyanonvara, Alicja~R Rudnicka, Christopher~G Owen, and Sarah~A Barman.
\newblock An ensemble classification-based approach applied to retinal blood vessel segmentation.
\newblock \emph{IEEE Transactions on Biomedical Engineering}, 59\penalty0 (9):\penalty0 2538--2548, 2012.

\bibitem[Gemma~Team et~al.(2024)Gemma~Team, Hardin, Dadashi, Bhupatiraju, Sifre, Rivière, Kale, Love, Tafti, Hussenot, and et~al.]{gemma_2024}
Thomas~Mesnard Gemma~Team, Cassidy Hardin, Robert Dadashi, Surya Bhupatiraju, Laurent Sifre, Morgane Rivière, Mihir~Sanjay Kale, Juliette Love, Pouya Tafti, Léonard Hussenot, and et al.
\newblock Gemma.
\newblock 2024.

\bibitem[Haug and Ostermann(2015)]{haug2015crop}
Sebastian Haug and J{\"o}rn Ostermann.
\newblock A crop/weed field image dataset for the evaluation of computer vision based precision agriculture tasks.
\newblock In \emph{Computer Vision-ECCV 2014 Workshops: Zurich, Switzerland, September 6-7 and 12, 2014, Proceedings, Part IV 13}, pages 105--116. Springer, 2015.

\bibitem[He et~al.(2024)He, Peng, Jiang, Hu, Zhang, Nie, Wang, and Wang]{he2024pointseg}
Qingdong He, Jinlong Peng, Zhengkai Jiang, Xiaobin Hu, Jiangning Zhang, Qiang Nie, Yabiao Wang, and Chengjie Wang.
\newblock Pointseg: A training-free paradigm for 3d scene segmentation via foundation models.
\newblock \emph{arXiv preprint arXiv:2403.06403}, 2024.

\bibitem[Hoyer et~al.(2022{\natexlab{a}})Hoyer, Dai, and Gool]{Hoyer22cvpr}
Lukas Hoyer, Dengxin Dai, and Luc~Van Gool.
\newblock Daformer: Improving network architectures and training strategies for domain-adaptive semantic segmentation.
\newblock In \emph{CVPR}, 2022{\natexlab{a}}.

\bibitem[Hoyer et~al.(2022{\natexlab{b}})Hoyer, Dai, and Gool]{Hoyer22eccv}
Lukas Hoyer, Dengxin Dai, and Luc~Van Gool.
\newblock Hrda: Context-aware high-resolution domain-adaptive semantic segmentation.
\newblock In \emph{ECCV}, 2022{\natexlab{b}}.

\bibitem[Hoyer et~al.(2023)Hoyer, Dai, Wang, and Gool]{Hoyer23}
Lukas Hoyer, Dengxin Dai, Haoran Wang, and Luc~Van Gool.
\newblock Mic: Masked image consistency for context-enhanced domain adaptation.
\newblock In \emph{CVPR}, 2023.

\bibitem[Islam et~al.(2020)Islam, Edge, Xiao, Luo, Mehtaz, Morse, Enan, and Sattar]{islam2020semantic}
Md~Jahidul Islam, Chelsey Edge, Yuyang Xiao, Peigen Luo, Muntaqim Mehtaz, Christopher Morse, Sadman~Sakib Enan, and Junaed Sattar.
\newblock Semantic segmentation of underwater imagery: Dataset and benchmark.
\newblock In \emph{2020 IEEE/RSJ International Conference on Intelligent Robots and Systems (IROS)}, pages 1769--1776. IEEE, 2020.

\bibitem[Jha et~al.(2021)Jha, Ali, Emanuelsen, Hicks, Thambawita, Garcia-Ceja, Riegler, de~Lange, Schmidt, Johansen, et~al.]{jha2021kvasir}
Debesh Jha, Sharib Ali, Krister Emanuelsen, Steven~A Hicks, Vajira Thambawita, Enrique Garcia-Ceja, Michael~A Riegler, Thomas de Lange, Peter~T Schmidt, H{\aa}vard~D Johansen, et~al.
\newblock Kvasir-instrument: Diagnostic and therapeutic tool segmentation dataset in gastrointestinal endoscopy.
\newblock In \emph{MultiMedia Modeling: 27th International Conference, MMM 2021, Prague, Czech Republic, June 22--24, 2021, Proceedings, Part II 27}, pages 218--229. Springer, 2021.

\bibitem[Jia et~al.(2021)Jia, Yang, Xia, Chen, Parekh, Pham, Le, Sung, Li, and Duerig]{align}
Chao Jia, Yinfei Yang, Ye Xia, Yi-Ting Chen, Zarana Parekh, Hieu Pham, Quoc Le, Yun-Hsuan Sung, Zhen Li, and Tom Duerig.
\newblock Scaling up visual and vision-language representation learning with noisy text supervision.
\newblock pages 4904--4916. PMLR, 2021.

\bibitem[Jia et~al.(2022)Jia, Tang, Chen, Cardie, Belongie, Hariharan, and Lim]{Jia2022VisualPT}
Menglin Jia, Luming Tang, Bor-Chun Chen, Claire Cardie, Serge~J. Belongie, Bharath Hariharan, and Ser~Nam Lim.
\newblock Visual prompt tuning.
\newblock \emph{ArXiv}, abs/2203.12119, 2022.

\bibitem[Jiang et~al.(2023)Jiang, Sablayrolles, Mensch, Bamford, Chaplot, Casas, Bressand, Lengyel, Lample, Saulnier, et~al.]{jiang2023mistral}
Albert~Q Jiang, Alexandre Sablayrolles, Arthur Mensch, Chris Bamford, Devendra~Singh Chaplot, Diego de~las Casas, Florian Bressand, Gianna Lengyel, Guillaume Lample, Lucile Saulnier, et~al.
\newblock Mistral 7b.
\newblock \emph{arXiv preprint arXiv:2310.06825}, 2023.

\bibitem[Jin et~al.(2024)Jin, Jiang, Huang, Lu, and Lu]{jin2024llms}
Sheng Jin, Xueying Jiang, Jiaxing Huang, Lewei Lu, and Shijian Lu.
\newblock Llms meet vlms: Boost open vocabulary object detection with fine-grained descriptors.
\newblock \emph{arXiv preprint arXiv:2402.04630}, 2024.

\bibitem[Kundu et~al.(2021)Kundu, Kulkarni, Singh, Jampani, and Babu]{Kundu21}
Jogendra~Nath Kundu, Akshay Kulkarni, Amit Singh, Varun Jampani, and R.~Venkatesh Babu.
\newblock Generalize then adapt: Source-free domain adaptive semantic segmentation.
\newblock In \emph{ICCV}, 2021.

\bibitem[Lan et~al.(2024)Lan, Chen, Ke, Wang, Feng, and Zhang]{proxyclip}
Mengcheng Lan, Chaofeng Chen, Yiping Ke, Xinjiang Wang, Litong Feng, and Wayne Zhang.
\newblock Proxyclip: Proxy attention improves clip for open-vocabulary segmentation.
\newblock In \emph{ECCV}, 2024.

\bibitem[Li et~al.(2020)Li, Kang, Liu, Wei, and Yang]{Li20}
Guangrui Li, Guoliang Kang, Wu Liu, Yunchao Wei, and Yi Yang.
\newblock Content-consistent matching for domain adaptive semantic segmentation.
\newblock In \emph{ECCV}, 2020.

\bibitem[Li et~al.(2017)Li, Zhao, Wei, Lang, Li, Sim, Yan, and Feng]{li2017multiple}
Jianshu Li, Jian Zhao, Yunchao Wei, Congyan Lang, Yidong Li, Terence Sim, Shuicheng Yan, and Jiashi Feng.
\newblock Multiple-human parsing in the wild.
\newblock \emph{arXiv preprint arXiv:1705.07206}, 2017.

\bibitem[Li et~al.(2023)Li, Savarese, and Hoi]{MUST}
Junnan Li, Silvio Savarese, and Steven C.~H. Hoi.
\newblock Masked unsupervised self-training for label-free image classification.
\newblock In \emph{ICLR}, 2023.

\bibitem[Liang et~al.(2023)Liang, Wu, Dai, Li, Zhao, Zhang, Zhang, Vajda, and Marculescu]{ovseg}
Feng Liang, Bichen Wu, Xiaoliang Dai, Kunpeng Li, Yinan Zhao, Hang Zhang, Peizhao Zhang, Peter Vajda, and Diana Marculescu.
\newblock Open-vocabulary semantic segmentation with mask-adapted clip.
\newblock In \emph{CVPR}, pages 7061--7070, 2023.

\bibitem[Liu et~al.(2019)Liu, Yao, Lu, Xie, and Li]{liu2019deepcrack}
Yahui Liu, Jian Yao, Xiaohu Lu, Renping Xie, and Li Li.
\newblock Deepcrack: A deep hierarchical feature learning architecture for crack segmentation.
\newblock \emph{Neurocomputing}, 338:\penalty0 139--153, 2019.

\bibitem[Lu et~al.(2022)Lu, Luo, Zhang, Li, Yang, and Xiao]{Lu22}
Yulei Lu, Yawei Luo, Li Zhang, Zheyang Li, Yi Yang, and Jun Xiao.
\newblock Bidirectional self-training with multiple anisotropic prototypes for domain adaptive semantic segmentation.
\newblock In \emph{ACM MM}, 2022.

\bibitem[Lyu et~al.(2020)Lyu, Vosselman, Xia, Yilmaz, and Yang]{lyu2020uavid}
Ye Lyu, George Vosselman, Gui-Song Xia, Alper Yilmaz, and Michael~Ying Yang.
\newblock Uavid: A semantic segmentation dataset for uav imagery.
\newblock \emph{ISPRS journal of photogrammetry and remote sensing}, 165:\penalty0 108--119, 2020.

\bibitem[Ma et~al.(2024)Ma, Zhang, Guo, and Xu]{ma2024swapprompt}
Xiaosong Ma, Jie Zhang, Song Guo, and Wenchao Xu.
\newblock Swapprompt: Test-time prompt adaptation for vision-language models.
\newblock \emph{NeurIPS}, 36, 2024.

\bibitem[Mahbod et~al.(2021)Mahbod, Schaefer, Bancher, L{\"o}w, Dorffner, Ecker, and Ellinger]{mahbod2021cryonuseg}
Amirreza Mahbod, Gerald Schaefer, Benjamin Bancher, Christine L{\"o}w, Georg Dorffner, Rupert Ecker, and Isabella Ellinger.
\newblock Cryonuseg: A dataset for nuclei instance segmentation of cryosectioned h\&e-stained histological images.
\newblock \emph{Computers in biology and medicine}, 132:\penalty0 104349, 2021.

\bibitem[Mata et~al.(2024)Mata, Ranasinghe, and Ryoo]{mata2024copt}
Cristina Mata, Kanchana Ranasinghe, and Michael Ryoo.
\newblock Copt: Unsupervised domain adaptive segmentation using domain-agnostic text embeddings.
\newblock In \emph{ECCV}, 2024.

\bibitem[Mateo-Garcia et~al.(2021)Mateo-Garcia, Veitch-Michaelis, Smith, Oprea, Schumann, Gal, Baydin, and Backes]{mateo2021towards}
Gonzalo Mateo-Garcia, Joshua Veitch-Michaelis, Lewis Smith, Silviu~Vlad Oprea, Guy Schumann, Yarin Gal, At{\i}l{\i}m~G{\"u}ne{\c{s}} Baydin, and Dietmar Backes.
\newblock Towards global flood mapping onboard low cost satellites with machine learning.
\newblock \emph{Scientific reports}, 11\penalty0 (1):\penalty0 7249, 2021.

\bibitem[Menon and Vondrick(2023)]{menon2022visual}
Sachit Menon and Carl Vondrick.
\newblock Visual classification via description from large language models.
\newblock \emph{ICLR}, 2023.

\bibitem[Ozturk et~al.(2024)Ozturk, Prabhushankar, and AlRegib]{ozturk2024intelligent}
Efe Ozturk, Mohit Prabhushankar, and Ghassan AlRegib.
\newblock Intelligent multi-view test time augmentation.
\newblock \emph{arXiv preprint arXiv:2406.08593}, 2024.

\bibitem[Radford et~al.(2021)Radford, Kim, Hallacy, Ramesh, Goh, Agarwal, Sastry, Askell, Mishkin, Clark, et~al.]{clip}
Alec Radford, Jong~Wook Kim, Chris Hallacy, Aditya Ramesh, Gabriel Goh, Sandhini Agarwal, Girish Sastry, Amanda Askell, Pamela Mishkin, Jack Clark, et~al.
\newblock Learning transferable visual models from natural language supervision.
\newblock pages 8748--8763. PMLR, 2021.

\bibitem[Rahnemoonfar et~al.(2021)Rahnemoonfar, Chowdhury, Sarkar, Varshney, Yari, and Murphy]{rahnemoonfar2021floodnet}
Maryam Rahnemoonfar, Tashnim Chowdhury, Argho Sarkar, Debvrat Varshney, Masoud Yari, and Robin~Roberson Murphy.
\newblock Floodnet: A high resolution aerial imagery dataset for post flood scene understanding.
\newblock \emph{IEEE Access}, 9:\penalty0 89644--89654, 2021.

\bibitem[Ranasinghe et~al.(2023)Ranasinghe, McKinzie, Ravi, Yang, Toshev, and Shlens]{clippy}
Kanchana Ranasinghe, Brandon McKinzie, Sachin Ravi, Yinfei Yang, Alexander Toshev, and Jonathon Shlens.
\newblock Perceptual grouping in contrastive vision-language models.
\newblock In \emph{CVPR}, pages 5571--5584, 2023.

\bibitem[Ren et~al.(2024)Ren, Liu, Zeng, Lin, Li, Cao, Chen, Huang, Chen, Yan, Zeng, Zhang, Li, Yang, Li, Jiang, and Zhang]{ren2024grounded}
Tianhe Ren, Shilong Liu, Ailing Zeng, Jing Lin, Kunchang Li, He Cao, Jiayu Chen, Xinyu Huang, Yukang Chen, Feng Yan, Zhaoyang Zeng, Hao Zhang, Feng Li, Jie Yang, Hongyang Li, Qing Jiang, and Lei Zhang.
\newblock Grounded sam: Assembling open-world models for diverse visual tasks, 2024.

\bibitem[Sakaridis et~al.(2019)Sakaridis, Dai, and Gool]{sakaridis2019guided}
Christos Sakaridis, Dengxin Dai, and Luc~Van Gool.
\newblock Guided curriculum model adaptation and uncertainty-aware evaluation for semantic nighttime image segmentation.
\newblock In \emph{ICCV}, pages 7374--7383, 2019.

\bibitem[Seibold et~al.(2022)Seibold, Rei{\ss}, Sarfraz, Fink, Mayer, Sellner, Kim, Maier-Hein, Kleesiek, and Stiefelhagen]{seibold2022detailed}
Constantin Seibold, Simon Rei{\ss}, Saquib Sarfraz, Matthias~A Fink, Victoria Mayer, Jan Sellner, Moon~Sung Kim, Klaus~H Maier-Hein, Jens Kleesiek, and Rainer Stiefelhagen.
\newblock Detailed annotations of chest x-rays via ct projection for report understanding.
\newblock \emph{arXiv preprint arXiv:2210.03416}, 2022.

\bibitem[Shang and Ryoo(2023)]{Shang2023ActiveVR}
Jinghuan Shang and Michael~S. Ryoo.
\newblock Active vision reinforcement learning under limited visual observability.
\newblock In \emph{NeurIPS}, 2023.

\bibitem[Shivakumar et~al.(2020)Shivakumar, Rodrigues, Zhou, Miller, Kumar, and Taylor]{shivakumar2020pst900}
Shreyas~S Shivakumar, Neil Rodrigues, Alex Zhou, Ian~D Miller, Vijay Kumar, and Camillo~J Taylor.
\newblock Pst900: Rgb-thermal calibration, dataset and segmentation network.
\newblock In \emph{2020 IEEE international conference on robotics and automation (ICRA)}, pages 9441--9447. IEEE, 2020.

\bibitem[Shu et~al.(2022)Shu, Nie, Huang, Yu, Goldstein, Anandkumar, and Xiao]{tpt}
Manli Shu, Weili Nie, De-An Huang, Zhiding Yu, Tom Goldstein, Anima Anandkumar, and Chaowei Xiao.
\newblock Test-time prompt tuning for zero-shot generalization in vision-language models.
\newblock \emph{NeurIPS}, 35:\penalty0 14274--14289, 2022.

\bibitem[Wah et~al.(2011{\natexlab{a}})Wah, Branson, Welinder, Perona, and Belongie]{WahCUB_200_2011}
C. Wah, S. Branson, P. Welinder, P. Perona, and S. Belongie.
\newblock Caltech-ucsd birds 200.
\newblock Technical Report CNS-TR-2011-001, California Institute of Technology, 2011{\natexlab{a}}.

\bibitem[Wah et~al.(2011{\natexlab{b}})Wah, Branson, Welinder, Perona, and Belongie]{welinder2010caltech}
Catherine Wah, Steve Branson, Peter Welinder, Pietro Perona, and Serge Belongie.
\newblock {Caltech-UCSD Birds 200}.
\newblock \emph{California Institute of Technology}, 2011{\natexlab{b}}.

\bibitem[Wang et~al.(2022)Wang, Liu, Suganuma, and Okatani]{Wang22}
Zhijie Wang, Xing Liu, Masanori Suganuma, and Takayuki Okatani.
\newblock Cross-region domain adaptation for class-level alignment, 2022.

\bibitem[Waqas~Zamir et~al.(2019)Waqas~Zamir, Arora, Gupta, Khan, Sun, Shahbaz~Khan, Zhu, Shao, Xia, and Bai]{waqas2019isaid}
Syed Waqas~Zamir, Aditya Arora, Akshita Gupta, Salman Khan, Guolei Sun, Fahad Shahbaz~Khan, Fan Zhu, Ling Shao, Gui-Song Xia, and Xiang Bai.
\newblock isaid: A large-scale dataset for instance segmentation in aerial images.
\newblock In \emph{CVPRW}, pages 28--37, 2019.

\bibitem[Wu et~al.(2021)Wu, Fu, Liu, Lim, Hoi, and Sun]{wu2021large}
Xiongwei Wu, Xin Fu, Ying Liu, Ee-Peng Lim, Steven~CH Hoi, and Qianru Sun.
\newblock A large-scale benchmark for food image segmentation.
\newblock In \emph{Proceedings of the 29th ACM international conference on multimedia}, pages 506--515, 2021.

\bibitem[Wysocza{\'n}ska et~al.(2023)Wysocza{\'n}ska, Sim{\'e}oni, Ramamonjisoa, Bursuc, Trzci{\'n}ski, and P{\'e}rez]{clip-dinoiser}
Monika Wysocza{\'n}ska, Oriane Sim{\'e}oni, Micha{\"e}l Ramamonjisoa, Andrei Bursuc, Tomasz Trzci{\'n}ski, and Patrick P{\'e}rez.
\newblock Clip-dinoiser: Teaching clip a few dino tricks for open-vocabulary semantic segmentation.
\newblock \emph{arXiv}, 2023.

\bibitem[Xu et~al.(2022)Xu, Zhang, Wei, Lin, Cao, Hu, and Bai]{xu2022simple}
Mengde Xu, Zheng Zhang, Fangyun Wei, Yutong Lin, Yue Cao, Han Hu, and Xiang Bai.
\newblock A simple baseline for open-vocabulary semantic segmentation with pre-trained vision-language model.
\newblock In \emph{ECCV}, pages 736--753. Springer, 2022.

\bibitem[Xu et~al.(2023)Xu, Zhang, Wei, Hu, and Bai]{xu2023side}
Mengde Xu, Zheng Zhang, Fangyun Wei, Han Hu, and Xiang Bai.
\newblock Side adapter network for open-vocabulary semantic segmentation.
\newblock In \emph{CVPR}, pages 2945--2954, 2023.

\bibitem[Yilmaz et~al.(2024)Yilmaz, Peng, Pollefeys, Engelmann, and Blum]{yilmaz2024opendas}
Gonca Yilmaz, Songyou Peng, Marc Pollefeys, Francis Engelmann, and Hermann Blum.
\newblock Opendas: Open-vocabulary domain adaptation for 2d and 3d segmentation.
\newblock \emph{arXiv preprint arXiv:2405.20141}, 2024.

\bibitem[Yu et~al.(2020{\natexlab{a}})Yu, Chen, Wang, Xian, Chen, Liu, Madhavan, and Darrell]{yu2020bdd100k}
Fisher Yu, Haofeng Chen, Xin Wang, Wenqi Xian, Yingying Chen, Fangchen Liu, Vashisht Madhavan, and Trevor Darrell.
\newblock Bdd100k: A diverse driving dataset for heterogeneous multitask learning.
\newblock In \emph{CVPR}, pages 2636--2645, 2020{\natexlab{a}}.

\bibitem[Yu et~al.(2024)Yu, Wang, Tu, Cao, Zhang-li, Lv, Peng, Yao, Zhang, Li, yan Li, Zhang, Bai, Liu, Xin, Lin, Yun, Gong, Chen, Wu, Qi, Li, Guan, Zeng, Qi, Jin, Liu, Gu, Gu, Yao, Ding, Hou, Liu, Xu, Tang, and Li]{Yu2023KoLACB}
Jifan Yu, Xiaozhi Wang, Shangqing Tu, Shulin Cao, Daniel Zhang-li, Xin Lv, Hao Peng, Zijun Yao, Xiaohan Zhang, Hanming Li, Chun yan Li, Zheyuan Zhang, Yushi Bai, Yantao Liu, Amy Xin, Nianyi Lin, Kaifeng Yun, Linlu Gong, Jianhui Chen, Zhili Wu, Yun~Peng Qi, Weikai Li, Yong Guan, Kaisheng Zeng, Ji Qi, Hailong Jin, Jinxin Liu, Yuxian Gu, Yu Gu, Yuan Yao, Ning Ding, Lei Hou, Zhiyuan Liu, Bin Xu, Jie Tang, and Juanzi Li.
\newblock Kola: Carefully benchmarking world knowledge of large language models.
\newblock In \emph{ICLR}, 2024.

\bibitem[Yu et~al.(2020{\natexlab{b}})Yu, Kumar, Gupta, Levine, Hausman, and Finn]{yu2020gradient}
Tianhe Yu, Saurabh Kumar, Abhishek Gupta, Sergey Levine, Karol Hausman, and Chelsea Finn.
\newblock Gradient surgery for multi-task learning.
\newblock \emph{NeurIPS}, 33:\penalty0 5824--5836, 2020{\natexlab{b}}.

\bibitem[Zhang et~al.(2023)Zhang, Li, Zou, Liu, Li, Yang, and Zhang]{zhang2023simple}
Hao Zhang, Feng Li, Xueyan Zou, Shilong Liu, Chunyuan Li, Jianwei Yang, and Lei Zhang.
\newblock A simple framework for open-vocabulary segmentation and detection.
\newblock In \emph{ICCV}, pages 1020--1031, 2023.

\bibitem[Zhang et~al.(2021)Zhang, Zhang, Zhang, Chen, Wang, and Wen]{Zhang21}
Pan Zhang, Bo Zhang, Ting Zhang, Dong Chen, Yong Wang, and Fang Wen.
\newblock Prototypical pseudo label denoising and target structure learning for domain adaptive semantic segmentation.
\newblock In \emph{CVPR}, 2021.

\bibitem[Zhao et~al.(2024)Zhao, Wang, Zhu, and Yang]{zhao2024testtime}
Shuai Zhao, Xiaohan Wang, Linchao Zhu, and Yi Yang.
\newblock Test-time adaptation with {CLIP} reward for zero-shot generalization in vision-language models.
\newblock In \emph{ICLR}, 2024.

\bibitem[Zheng and Yang(2020)]{Zheng20}
Zhedong Zheng and Yi Yang.
\newblock Unsupervised scene adaptation with memory regularization in vivo.
\newblock In \emph{IJCAI}, 2020.

\bibitem[Zhou et~al.(2022{\natexlab{a}})Zhou, Yang, Loy, and Liu]{zhou2022cocoop}
Kaiyang Zhou, Jingkang Yang, Chen~Change Loy, and Ziwei Liu.
\newblock Conditional prompt learning for vision-language models.
\newblock In \emph{CVPR}, 2022{\natexlab{a}}.

\bibitem[Zhou et~al.(2022{\natexlab{b}})Zhou, Yang, Loy, and Liu]{zhou2022coop}
Kaiyang Zhou, Jingkang Yang, Chen~Change Loy, and Ziwei Liu.
\newblock Learning to prompt for vision-language models.
\newblock \emph{IJCV}, 2022{\natexlab{b}}.

\bibitem[Zou et~al.(2023)Zou, Dou, Yang, Gan, Li, Li, Dai, Behl, Wang, Yuan, et~al.]{zou2023generalized}
Xueyan Zou, Zi-Yi Dou, Jianwei Yang, Zhe Gan, Linjie Li, Chunyuan Li, Xiyang Dai, Harkirat Behl, Jianfeng Wang, Lu Yuan, et~al.
\newblock Generalized decoding for pixel, image, and language.
\newblock In \emph{CVPR}, pages 15116--15127, 2023.

\end{thebibliography}
}

\clearpage
\setcounter{page}{1}
\maketitlesupplementary

\appendix

\section{More Details}

\subsection{LLM based Category Attribute Generation}
\label{app:cat_attributes}
In this section, we provide an in-depth overview of the methods and strategies we used to generate visually descriptive attributes for each object category within the Open Vocabulary Semantic Segmentation (OVSS) task.

\subsubsection{Selection of Large Language Models}
The quality of the visual attributes employed in our method significantly influences performance, as demonstrated in \Cref{tbl:llm_attributes}. This evaluation highlights the performance impact of different attribute sets generated by three different large language models (LLMs), emphasizing the importance of selecting high-quality attributes for optimal results.

The quality of the attributes is highly correlated with the quality of the LLM. To identify the most suitable LLM for OVSS, we evaluate several open-source LLMs. The selection process prioritizes models capable of accurately and reliably following user instructions, a critical requirement for generating well-structured and relevant attributes. Open-source models are preferred due to their accessibility, transparency, and flexibility, which enable effective customization for task-specific needs.

Among the evaluated models, the Llama 3 Instruct 70B \cite{llama3modelcard}, a fine-tuned variant optimized for instruction-following tasks, demonstrates superior performance. Additionally, we explore the 2B Instruct variant of the Gemma model \cite{gemma_2024} and the instruction-tuned Mistral-7B-v0.2 model \cite{jiang2023mistral}. We observe a positive correlation between model size, in terms of parameter count, and task performance, aligning with established expectations. Furthermore, instruction-tuned models consistently exhibit enhanced adaptability, reliably generating outputs in the desired format and confirming their effectiveness in user-guided attribute generation.


\begin{table}[h]
\centering
\small
\def\arraystretch{1.2}  
\setlength\tabcolsep{0.8em}  
\scalebox{0.95}{
\begin{tabular}{c|l}
\toprule
LLM           & DZ    \\ \midrule
Gemma-2B      & 33.4  \\
Mistral-7B    & 33.2  \\
\rowcolor{Gray}
Llama3-70B    & 34.6  \\ \bottomrule
\end{tabular}
}
\caption{\footnotesize
\textbf{Selection of LLM}: 
We report mIoU (\%) on Dark Zurich (DZ) \cite{sakaridis2019guided} dataset for attributes generated by Gemma-2B-Instruct (Gemma-2B) \cite{gemma_2024}, Mistral-7B-Instruct-v0.2 (Mistral-7B) \cite{jiang2023mistral} and Meta-Llama-3-70B-Instruct (Llama3-70B) \cite{llama3modelcard} LLMs.
}
\label{tbl:llm_attributes}
\vspace{-1em}
\end{table}
\begin{figure}[h]
    \centering
    \includegraphics[width=1.0\linewidth]
    {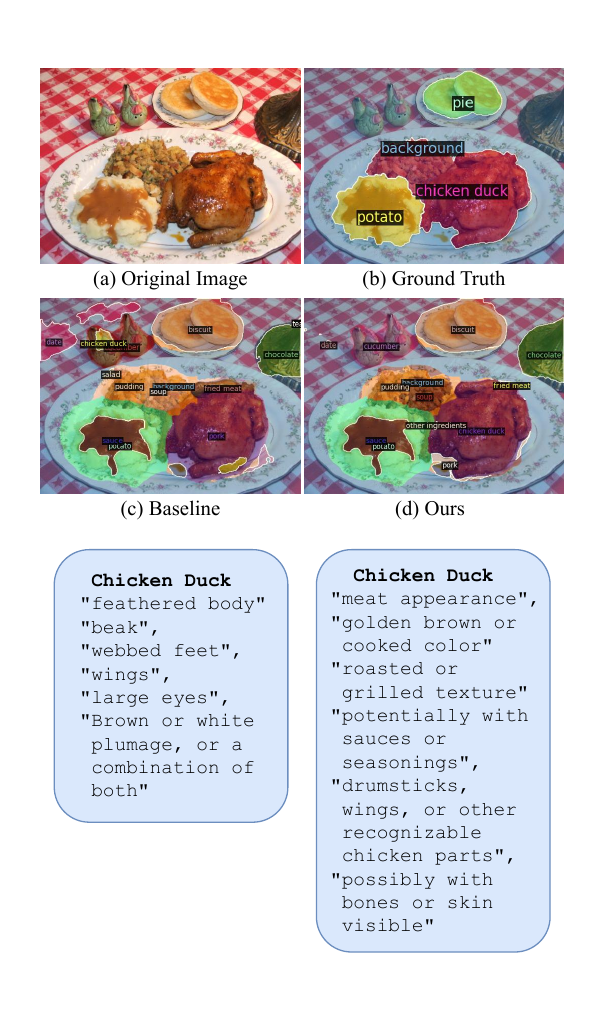}
    \caption{\textbf{Illustration of improved attribute generation for FoodSeg103\cite{wu2021large} dataset images} (a) The original image. (b) Ground truth segmentation map. (c) Baseline \cite{menon2022visual} attribute generation method, which included general and irrelevant features such as ``feathered body'' and ``wings'' for ``chicken duck.'' (d) Our approach with dataset-specific descriptions (e.g., ``photo of food''), resulting in more relevant attributes like ``roasted or grilled texture'' and ``golden brown or cooked color.''
    }
    \label{fig:attributes}
\end{figure}

\begin{table*}[h]
\centering
\small
\def\arraystretch{1.2}  
\setlength\tabcolsep{0.8em}  
\scalebox{0.95}{
\begin{tabular}{lll}
\toprule
Dataset & Image type & Category descriptions \\ 
\midrule \midrule
BDD100K\cite{yu2020bdd100k} & `photo'  &  -  \\ 
Dark Zurich\cite{sakaridis2019guided} & `photo' &  `background' = `background of a photo taken while driving at night' \\
MHP v1\cite{li2017multiple} & `photo'  &  - \\ 
FoodSeg103\cite{wu2021large} & `photo of food'  &  `background' = `background of food' \\
ATLANTIS\cite{erfani2022atlantis} & `photo'  &  - \\ 
DRAM\cite{cohen2022semantic} & `photo'  &  - \\ \midrule
iSAID\cite{waqas2019isaid} & `aerial image'  & `background' = `background of aerial images' \\ 
ISPRS Pots.\cite{alemohammad2020landcovernet} & `IRRG color map aerial image'  &  - \\ 
WorldFloods\cite{mateo2021towards} & `IRRG color map aerial image'  &  - \\ 
FloodNet\cite{rahnemoonfar2021floodnet} & `photo'  &  - \\
UAVid\cite{lyu2020uavid} & `photo'  &  - \\  \midrule
Kvasir-Inst.\cite{jha2021kvasir} & `photo'  &  `others' = `gastrointestinal (GI) tract tissue', `tool' = `endoscopic grasping tool' \\ 
CHASE DB1\cite{fraz2012ensemble} & `photo'  &  `background' = `background of blood vessels in a retinal image' \\ 
CryoNuSeg\cite{mahbod2021cryonuseg} & `photo'  & `background' = `background of nuclei on a slide' \\
PAXRay-4\cite{seibold2022detailed} & `x-ray image'  &  - \\  \midrule
Corrosion CS\cite{bianchi2021corrosion} & `photo'  &  `others' = `regions such as the concrete surfaces, metal surfaces or environment' \\ 
DeepCrack\cite{liu2019deepcrack} & `photo'  &  - \\ 
PST900\cite{shivakumar2020pst900} & `thermal image'  &  - \\
ZeroWaste-f\cite{bashkirova2022zerowaste} & `conveyor belt image'  &  - \\  \midrule
SUIM\cite{islam2020semantic} & `photo'  &  - \\ 
CUB-200\cite{welinder2010caltech} & `photo'  &  - \\ 
CWFID\cite{haug2015crop} & `photo'  &  - \\
\bottomrule
\end{tabular}
}
\caption{
\textbf{Prompting techniques} In the prompt described in section \ref{app:attr_quality}, the original \texttt{category name} is substituted with the corresponding category description, and the \texttt{image type} is replaced with the specified image type provided in the table.}
\label{tbl:llm_prmpts}
\vspace{-1em}
\end{table*}

\subsubsection{Prompting Styles and Techniques}
\label{app:attr_quality}
\noindent\begin{minipage}[t]{0.45\textwidth}
\small
\vspace{0.75em} 
\begin{verbatim}
Q: What are useful visual attributes for 
   distinguishing a {category name} 
   from {','.join(other categories except 
   category name)} in a {image type}?
A: There are several useful visual 
   attributes to tell there is a 
   {category name} in a {image type}:
-
\end{verbatim}
\end{minipage}\\

We experimented with several prompts and ultimately adopted the above one, inspired by \cite{menon2022visual}, which was originally designed for attribute generation in classification tasks. In segmentation, however, multiple categories need to be identified within a single image, so the attributes must effectively distinguish each category from the others. To achieve this, we add a component, listing all category names in the prompt, allowing the LLM to identify which categories to distinguish from the given category. This approach helps ensure that the generated attributes effectively differentiate the target category from the other specified categories.

To further assist the LLM in generating relevant attributes, we provide specific descriptions of image types for certain datasets. For instance, labelling the \texttt{image type} as ``photo of food'' for the FoodSeg103 \cite{wu2021large} dataset prevents the LLM from producing more general or irrelevant attributes for category names (see \Cref{fig:attributes}). For other datasets, we simply specify the \texttt{image type} as ``photo''.  Additionally, for categories where the name alone is insufficiently descriptive (e.g., ``background", ``others'', ``tool''), we include a brief description to help the LLM generate relevant attributes. A comprehensive overview of these prompting techniques is provided in table \ref{tbl:llm_prmpts}.



\begin{table*}
    \begin{center}
       
    \resizebox{\textwidth}{!}{
    \begin{tabular}{llccl}
\toprule
Dataset & Task  & \# of categories & Number of Images & Categories \\
\midrule\midrule
BDD100K \cite{yu2020bdd100k} & Driving & 19 & 1,000 & [road; sidewalk; building; wall; fence; pole; traffic light; traffic sign; ...] \\
Dark Zurich \cite{sakaridis2019guided} & Driving & 20 & 50 & [unlabeled; road; sidewalk; building; wall; fence; pole; traffic light; ...] \\
MHP v1 \cite{li2017multiple} & Body parts & 19 & 980 & [others; hat; hair; sunglasses; upper clothes; skirt; pants; dress; ...] \\
FoodSeg103 \cite{wu2021large} & Ingredients & 104 & 2135 & [background; candy; egg tart; french fries; chocolate; biscuit; popcorn; ...] \\
ATLANTIS \cite{erfani2022atlantis} & Maritime & 56 & 1295 & [bicycle; boat; breakwater; bridge; building; bus; canal; car; ...] \\
DRAM \cite{cohen2022semantic} & Paintings & 12 & 718 & [bird; boat; bottle; cat; chair; cow; dog; horse; ...] \\
\hline
iSAID \cite{waqas2019isaid} & Objects & 16 & 4055 & [others; boat; storage tank; baseball diamond; tennis court; bridge; ...] \\
ISPRS Potsdam \cite{alemohammad2020landcovernet} & Land Use & 6 & 504 & [road; building; grass; tree; car; others] \\
WorldFloods \cite{mateo2021towards} & Floods & 3 & 160 & [land; water and flood; cloud] \\
FloodNet \cite{rahnemoonfar2021floodnet} & Floods  & 10 & 5571 & [building-flooded; building-non-flooded; road-flooded; water; tree; ...] \\
UAVid \cite{lyu2020uavid} & Objects & 8 & 840 & [others; building; road; tree; grass; moving car; parked car; humans] \\
\hline
Kvasir-Inst. \cite{jha2021kvasir} & Endoscopy & 2 & 118 & [others; tool] \\
CHASE DB1 \cite{fraz2012ensemble} & Retina Scan & 2 & 20 & [others; blood vessels] \\
CryoNuSeg \cite{mahbod2021cryonuseg} & WSI & 2 & 30 & [others; nuclei in cells] \\
PAXRay-4 \cite{seibold2022detailed} & X-Ray & 4x2 & 180 & [others, lungs], [others, bones], [others, mediastinum], [others, diaphragm] \\
\hline
Corrosion CS \cite{bianchi2021corrosion} & Corrosion & 4 & 44 & [others; steel with fair corrosion; ... poor corrosion; ... severe corrosion] \\
DeepCrack \cite{liu2019deepcrack} & Cracks & 2 & 237 & [concrete or asphalt; crack] \\
PST900 \cite{shivakumar2020pst900} & Coveryor & 5 & 929 & [background; fire extinguisher; backpack; drill; human] \\
ZeroWaste-f \cite{bashkirova2022zerowaste} & Thermal & 5 & 288 & [background or trash; rigid plastic; cardboard; metal; soft plastic] \\
\hline
SUIM \cite{islam2020semantic} & Underwater & 8 & 110 & [human diver; reefs and invertebrates; fish and vertebrates; ...] \\
CUB-200 \cite{welinder2010caltech}  & Bird species & 201 & 5794 & [background; Laysan Albatross; Sooty Albatross; Crested Auklet; ...] \\
CWFID \cite{haug2015crop} & Crops & 3 & 21 & [ground; crop seedling; weed] \\
\bottomrule
    \end{tabular}
}
    \end{center}
    \vspace{-10pt}
        \caption{\textbf{Details of the datasets in the MESS benchmark~\cite{MESS}} 
    }\label{tab:mess-detail}
\end{table*}

\subsection{Additional Details on Attribute Aggregation}
\label{app:attr_aggr}
\noindent\textbf{Attribute Aggregation in CAT-Seg} In CAT-Seg \cite{catseg}, the dimension of the prompt templates must remain fixed to pass through the Aggregator component. Therefore, rather than averaging across \( p \) prompts, as described in equation \ref{eq:final_txt_embd}, we use the concatenation of \( \{b^j_k \mid k \in [1, p] \} \) (see section \ref{subsection:ttfo}) with \( \{z^j_k \mid k \in [1, 80-p] \} \), where the \( 80-p \) non-learnable prompts for each category \( j \) come from the ImageNet templates used in CAT-Seg \cite{catseg}. For attributes, we utilize all 80 ImageNet templates employed in the CAT-Seg \cite{catseg}, denoted as \( \{\gamma_\text{{attr}}(A^j_k) \mid k \in [1, 80] \} \).

To obtain the final text embedding for each category $j$ for a given image $\rmX$, 
\begin{equation}
    \begin{aligned}
        \rvf_t^j &= \beta \big(\{b^j_k \mid k \in [1, p] \} \|  \{z^j_k \mid k \in [1, 80-p] \} \big) \\
        &\quad + (1-\beta) \{ \gamma_{\text{attr}}(A^j_k) \mid k \in [1, 80] \}
    \end{aligned}
\end{equation}
where $\beta$ is a hyper-parameter which we fix experimentally and $\|$ denotes concatenation operation. We obtain embeddings for all $n$ categories and 80 prompts as $\rmF_t = [\rvf_t^1, \rvf_t^2, ... , \rvf_t^n]$ the final text embeddings for the given image $\rmX$.


\subsection{Additional Details on Visual Aggregation}
\label{app:vis_aggr}
For TTFO, we observe a significant effect from cross-entropy loss but for selection, the effect is minimized. In TTFO we are tuning the prompts based on the loss values. However, we only use loss to sort the augmentations in selection. We assume that is the reason for the low effect on selection. Therefore, although we use \Cref{eq:ssl} for TTFO (\Cref{subsection:ttfo}), we modify it in augmentation selection (\Cref{subsec:vis_aggr}) as follows.
\begin{align}
    \gL_{\text{SSL-Augs}}^q &= \gamma_{\text{sel}} \left( \{ \gL_{\text{ent}}^{q, i} (\sF_v,\sF_{t,j}) \mid i \in [1, m] \}
    \right) \label{app:eq:ssl2} \\
    \gL_{\text{SSL-Augs}} &= 
    \gamma_{\text{aggr}} \left( \{ \gL_{\text{SSL-Augs}}^{q} \mid q \in \sR^{h' \times w'} \} \right) 
\label{app:eq:ssl}
\end{align}

\begin{table}[h]
\centering
\small
\def\arraystretch{1.2}  
\setlength\tabcolsep{0.8em}  
\scalebox{0.95}{
\begin{tabular}{l|c}
\toprule
Aug. Select Method  &  DZ   
\\ 
\midrule
$\gL_{\text{SSL}}^q$ (Eq: \ref{eq:ssl}) &  34.58 
\\
$\gL_{\text{SSL-Augs}}^q$ (Eq: \ref{app:eq:ssl}) &  34.55  
\\  
\bottomrule
\end{tabular}
}
\caption{
\textbf{Results under different augmentation selection loss functions: } We observe no significant changes in results with or without cross-entropy loss in augmentation selection.
}
\label{tbl:attr_aggregation}
\vspace{-1em}
\end{table}

\subsection{Dataset Details and Examples}
\label{app:dataset}
We thoroughly evaluate the MESS \cite{MESS} benchmark. It consists of 22 datasets from domains such as engineering, medical sciences, earth monitoring, agriculture, and biology. Additionally, the benchmark includes six datasets from diverse general classes including body parts, ingredients, paintings, maritime and driving. The benchmark consists of two datasets each taken from microscopic sensors, three datasets from electromagnetic sensors and others from visible spectrum sensors. There are datasets such as corrosion-cs \cite{bianchi2021corrosion} and zerowaste-f \cite{bucher2019zero} with a high-category similarity. The segment sizes vary from small to medium to large. The category vocabulary ranges from generic to task- and domain-specific. We refer the reader to MESS \cite{MESS} paper for more details and \Cref{tab:mess-detail} for additional dataset details.

\subsection{Details on Baselines}
\label{app:baselines}
We choose two CAT-Seg \cite{catseg} variants and CLIP-DINOiser \cite{clip-dinoiser} as baselines for evaluating our framework. They represent SOTA in their respective supervised and self-supervised approaches.\\
\noindent \textbf{Implementation of VFA in CAT-Seg:} CAT-Seg \cite{catseg} processes an image by diving it into overlapping patches. For each patch and the original image, two types of visual features are considered: (1) visual features from the backbone network and (2) visual features from CLIP's \cite{clip} visual encoder. In VFA, We update the original image's visual features (both backbone and clip features) using corresponding filtered crop features as described in \Cref{eq:vfa}. For the patches, we update visual features (from both backbone and CLIP) only if the filtered crop lies within the spatial region of the patch.

\noindent\textbf{Implementation of VFA in CLIP-DINOiser:} In CLIP-DINOiser \cite{clip-dinoiser}, we adapt VFA to update DINOised features. Specifically, we update the DINOised features of the original image using the DINOised features of the filtered crops. The updating process is as discussed in \Cref{eq:vfa}.

We forward the reader to CAT-Seg \cite{catseg} and CLIP-DINOiser \cite{clip-dinoiser} works for their exact architecture.

\subsection{Additional Ablations}
\label{app:additional_ablations}
\noindent\textbf{Effect of the loss function:} We use a combination of entropy minimization and a pseudo-labeling-based cross-entropy loss.
We ablate in Table \ref{tbl:spatial_aggr} the performance of different patch entropy aggregation methods in entropy minimization. We take the mean of all patches for calculation. However, to improve spatial awareness of the loss function we incorporate cross-entropy loss which takes into account \textit{good} patch-wise predictions. According to the results in Table \ref{tbl:loss_fn}, we establish the effectiveness of our loss function. 

\noindent\textbf{Learnable component in TTO for the textual modality:} As shown in \Cref{ablate:tpt_learneable_component}, tuning both prompt and per-class embeddings (PCE) leads to a significant improvement in performance over single-component tuning. We hypothesize that this improvement results from the synergistic roles of the two embeddings: while prompt embeddings enhance general adaptability to out-of-domain (OOD) data, per-class embeddings refine category-specific representations, that may not be well represented in the pre-trained general category embeddings.

\noindent\textbf{Attribute Aggregation:} We analyze influence of attribute aggregation on segmentation performance in \Cref{ablate:attr_aggregation}. \\ 
\textbf{(1) Test time attribute tuning:} We tune the attributes at test time, by treating attributes as an additional set of category names, which substantially increases memory consumption due to the multiplied category count by the attribute count per category. We then calculate the maximum probability between the category name with prompts and either the maximum or mean probability of the relevant attributes. This approach, with a loss function that maximizes one category name per patch, either emphasizes the relevant category name or one of its attributes. \\ 
However, attribute tuning is highly sensitive to the attribute set, leading to potential variations of ±10\% in mIoU. We hypothesize that this sensitivity arises from treating attributes as additional category names. In contrast, our method tunes only the prompts and category names, making it more robust to variations in the LLM-generated attribute set.\\
\textbf{(2) Post-aggregation:} This method is similar to the previous one but without the tuning process, still treating attributes as additional category names.  \\ 
\textbf{(3) Pre-aggregation:} This method is detailed in section \ref{subsec:text_attr}.\\
The presence of similar attributes across categories can cause ambiguity, as the model may struggle to distinguish whether an input corresponds to the feature of one category or another, affecting both attribute tuning and post-aggregation. Consequently, we select pre-aggregation as the optimal method, as it minimizes the influence of low-quality attributes while maintaining performance.


\begin{table}[t]
\centering
\small
\def\arraystretch{1.2}  
\setlength\tabcolsep{0.6em}  
\scalebox{0.85}{
\begin{tabular}{@{}ll|c@{}}
\toprule
& Loss function   &   DZ   \\ \midrule
\textbf{(I)} & Entropy minimization    &  34.4 \\  
\textbf{(II)} & Cross entropy           &  34.5 \\ \rowcolor{Gray} 
\textbf{(III)} & \textbf{(I)} + \textbf{(II)}     &  34.6 \\  \bottomrule
\end{tabular}
}
\caption{\footnotesize
\textbf{Results under different loss functions}: Pseudo-labeling based cross-entropy loss function improves the results over using entropy minimization on its own.
}
\label{tbl:loss_fn}
\end{table}

\begin{table}[t]
\centering
\small
\def\arraystretch{1.2}  
\setlength\tabcolsep{0.6em}  
\scalebox{0.85}{
\begin{tabular}{l|c}
\toprule
Method &   DZ  \\ \midrule 
Max   & 34.12 \\ 
Median    & 34.56 \\ \rowcolor{Gray} 
Mean     &  34.58 \\   \bottomrule
\end{tabular}
}
\caption{\footnotesize
\textbf{Spatial Aggregation}: We ablate maximum, median, and mean spatial aggregation and report mIoU (\%) on Dark Zurich dataset.  
}
\label{tbl:spatial_aggr}
\end{table}


\begin{SCfigure*}[][t]
    \centering
    \caption{\textbf{Qualitative comparison between Vis. Feat. Aggr. and Test Time Opt.:} Our approach (d) successfully identifies more \textit{fish} and (e) identifies \textit{sea-floor}, whereas baseline (c) fails.}
    \includegraphics[width=0.75\textwidth]{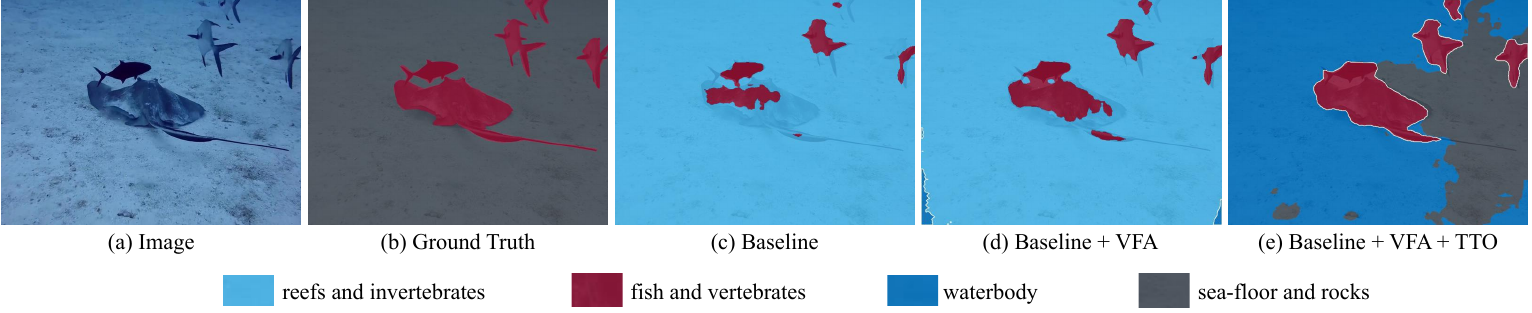}
    \label{fig:analysis}
    \vspace{1em}
\end{SCfigure*}
\begin{figure*}[t]
    \centering
    \includegraphics[width=\textwidth, keepaspectratio]{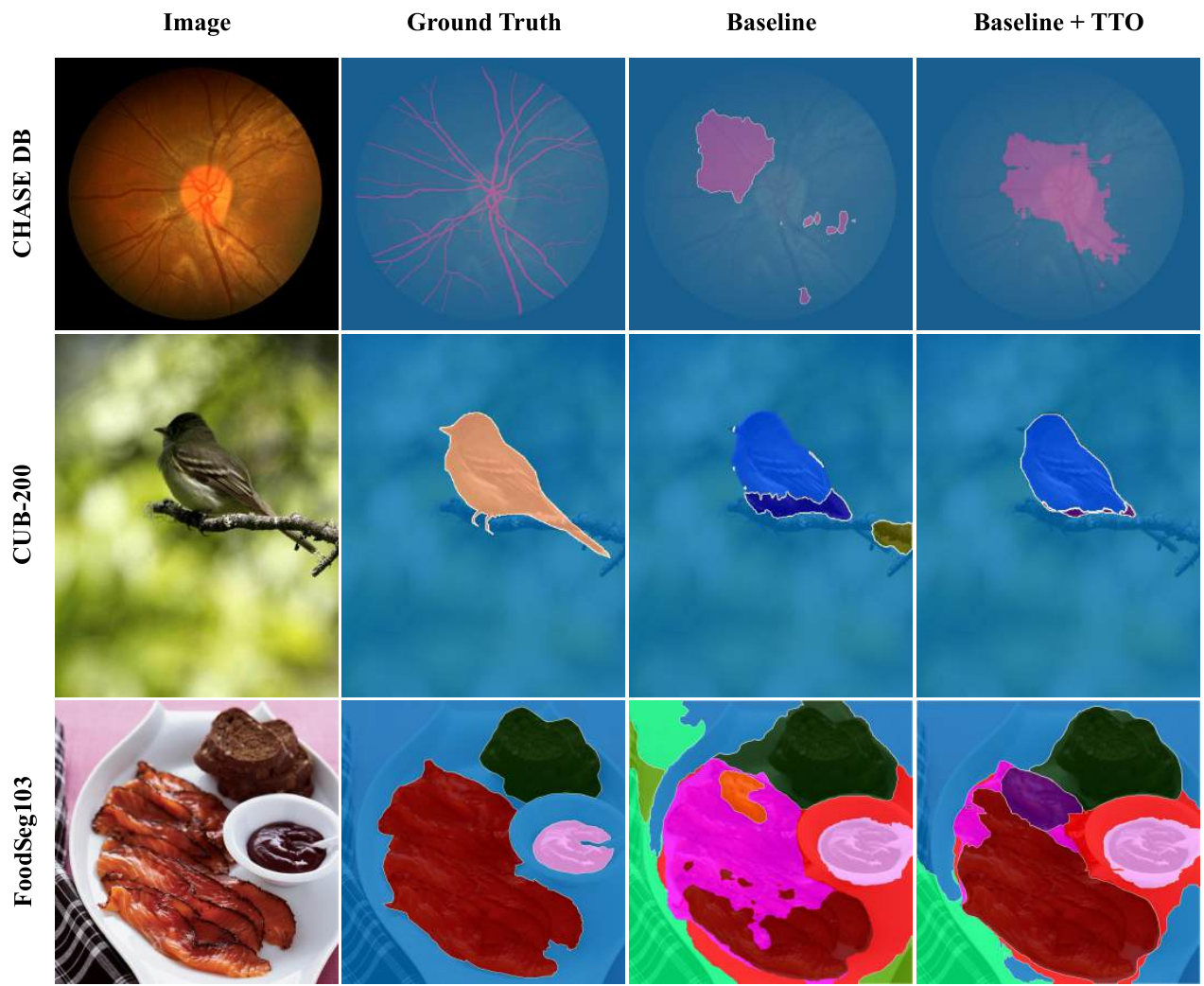}
    \caption{\textbf{Qualitative Evaluation:}
    We illustrate both success and failure cases of our proposed \modelname. We highlight how \modelname is still better than the baseline even in failure cases.
    }
    \label{app:fig:images}
\end{figure*}

\begin{figure*}[t]
    \centering
    \includegraphics[width=1.0\textwidth]{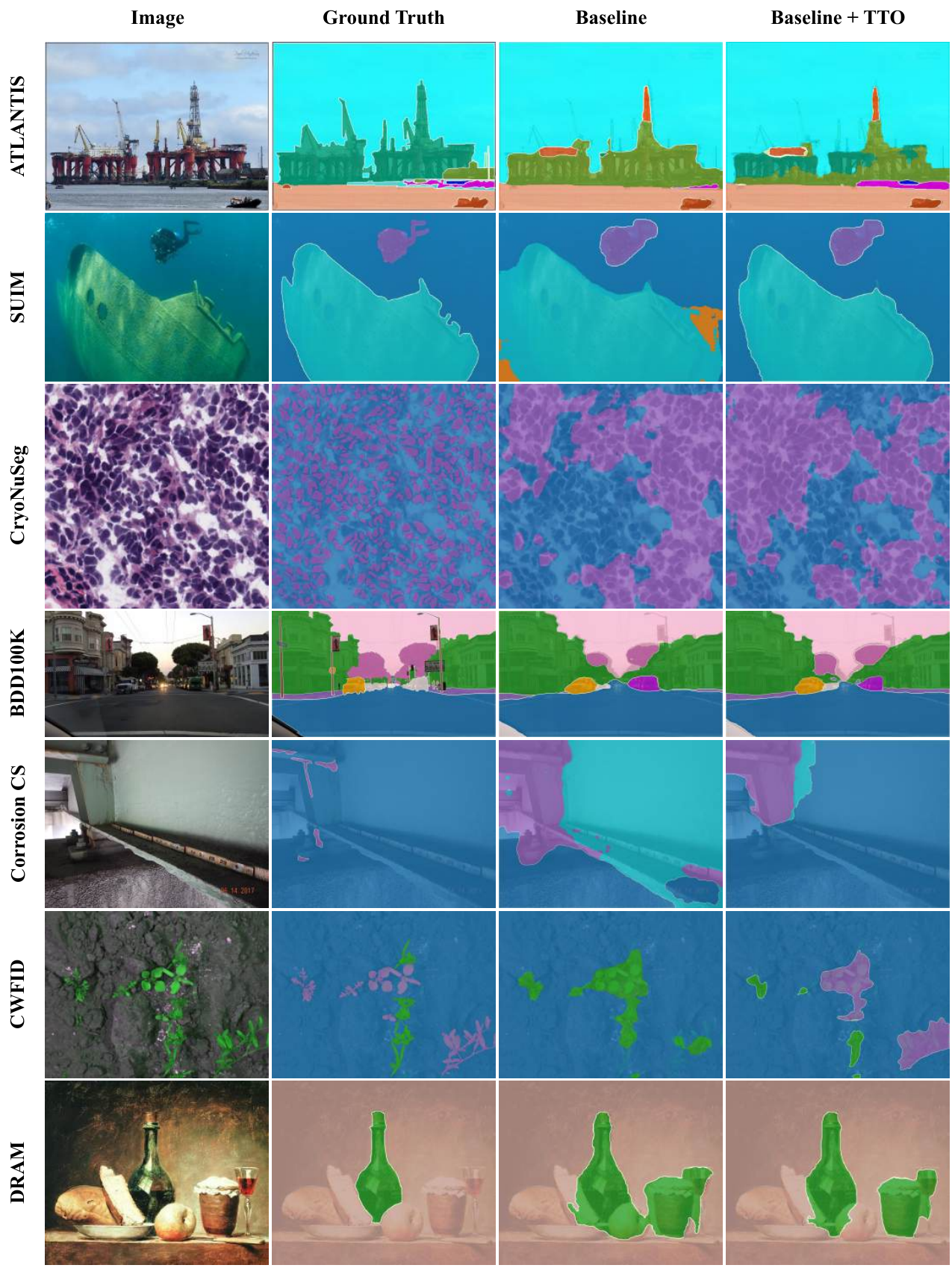}
    \caption{\textbf{Qualitative Evaluation:}
    We illustrate both success and failure cases of our proposed \modelname. We highlight how \modelname is still better than the baseline even in failure cases.
    }
    \label{app:fig:images}
\end{figure*}
\begin{figure*}[t]
    \centering
    \includegraphics[width=\textwidth,height=\textheight, keepaspectratio]{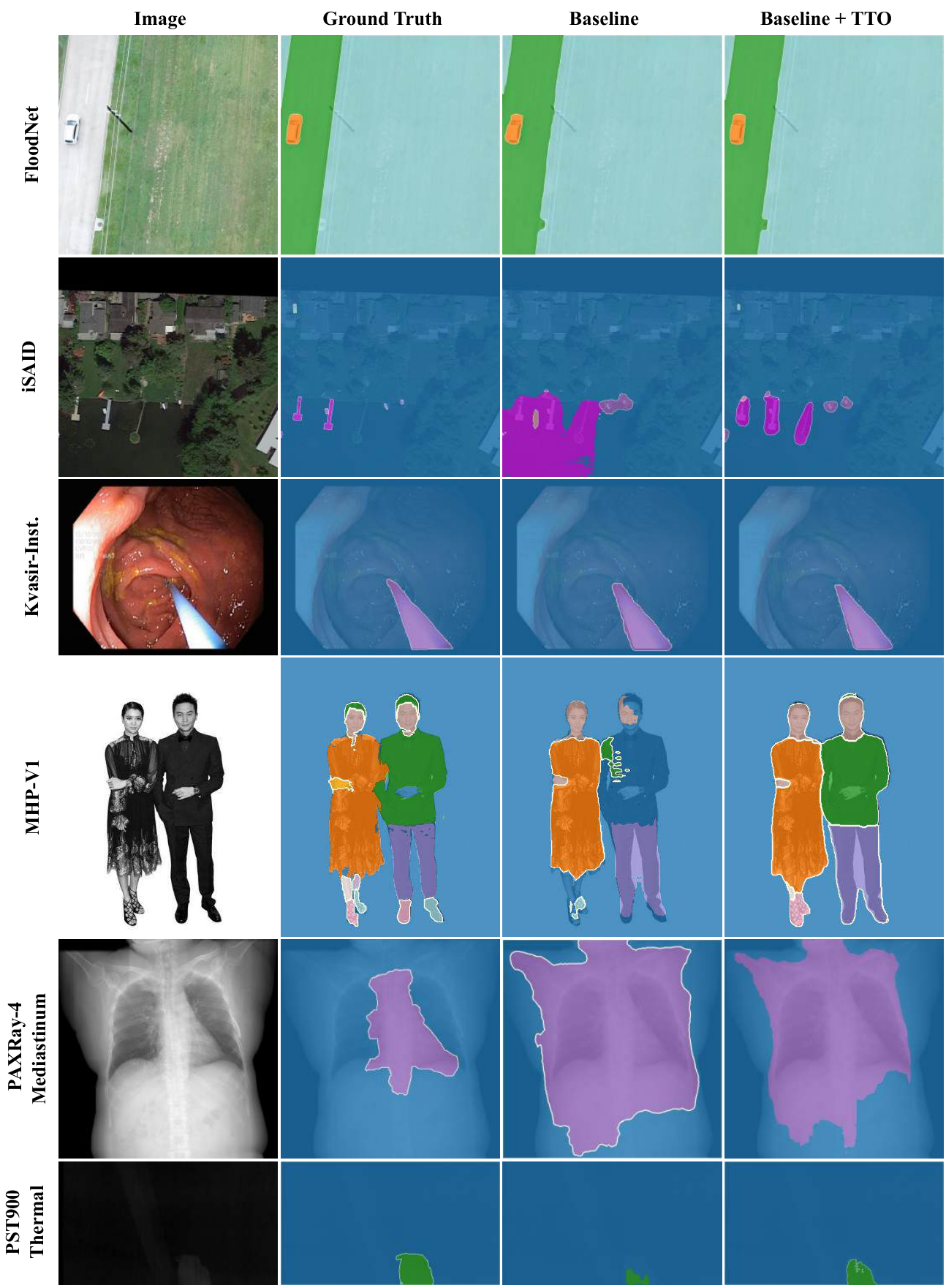}
    \caption{\textbf{Qualitative Evaluation:}
    We illustrate both success and failure cases of our proposed \modelname. We highlight how \modelname is still better than the baseline even in failure cases.
    }
    \label{app:fig:images}
\end{figure*}
\begin{figure*}[t]
    \centering
    \includegraphics[width=\textwidth,height=\textheight, keepaspectratio]{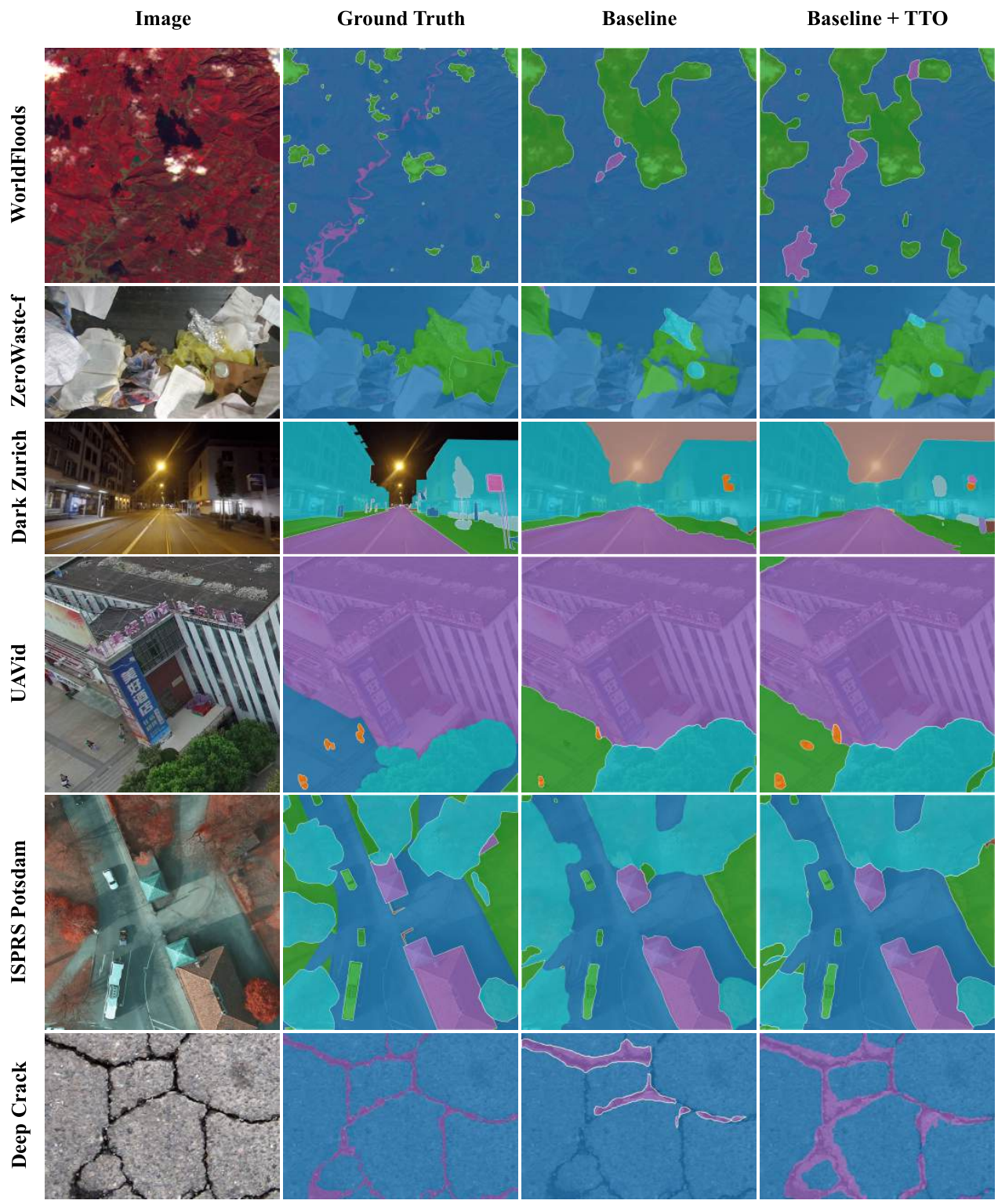}
    \caption{\textbf{Qualitative Evaluation:}
    We illustrate both success and failure cases of our proposed \modelname. We highlight how \modelname is still better than the baseline even in failure cases.
    }
    \label{app:fig:images}
\end{figure*}




\end{document}